\pdfoutput=1
\documentclass[10pt,letterpaper]{article}
\usepackage[top=0.85in,left=2.75in,footskip=0.75in]{geometry}

\usepackage{amsmath,amssymb}

\usepackage{changepage}

\usepackage{textcomp,marvosym}

\usepackage{cite}

\usepackage{nameref,hyperref}

\usepackage[right]{lineno}

\usepackage[nopatch=eqnum]{microtype}
\DisableLigatures[f]{encoding = *, family = * }

\usepackage[table]{xcolor}

\usepackage{array}

\newcolumntype{+}{!{\vrule width 2pt}}

\newlength\savedwidth

\raggedright
\usepackage[aboveskip=1pt,labelfont=bf,labelsep=period,justification=raggedright,singlelinecheck=off]{caption}

\makeatletter
\renewcommand{\@biblabel}[1]{\quad#1.}
\makeatother

\usepackage{lastpage,fancyhdr,graphicx}
\usepackage{epstopdf}
\fancyheadoffset[L]{2.25in}
\fancyfootoffset[L]{2.25in}
\usepackage{amsthm}
\usepackage{booktabs,tabularx}
\usepackage{enumitem}
\usepackage{algorithm}
\usepackage[noend]{algpseudocode}
\usepackage{url}
\usepackage[capitalise,nameinlink]{cleveref}
\crefname{figure}{Fig}{Figs}
\Crefname{figure}{Fig}{Figs}
\crefname{table}{Table}{Tables}
\Crefname{table}{Table}{Tables}
\crefname{algorithm}{Algorithm}{Algorithms}
\crefformat{equation}{Eq~(#2#1#3)}
\Crefformat{equation}{Eq~(#2#1#3)}
\crefmultiformat{equation}{Eqs~(#2#1#3)}{ and~(#2#1#3)}{, (#2#1#3)}{ and~(#2#1#3)}
\Crefmultiformat{equation}{Eqs~(#2#1#3)}{ and~(#2#1#3)}{, (#2#1#3)}{ and~(#2#1#3)}
\newtheorem{definition}{Definition}
\newtheorem{remark}{Remark}
\crefname{remark}{Remark}{Remarks}
\crefname{definition}{Definition}{Definitions}

\newcommand{\nreg}{\mathrm{nreg}}
\newcommand{\AUC}{\mathrm{AUC}}
\newcommand{\TTT}{\mathrm{TTT}_{0.05}}
\newcommand{\pool}{\mathcal{A}}
\newcommand{\lib}{\mathcal{L}}
\newcommand{\ns}{\nu}
\newcommand{\EI}{\mathrm{EI}}
\hypersetup{hidelinks}

\begin{document}
\vspace*{0.2in}

\begin{flushleft}
{\Large
\textbf{QuaNAS: Uncertainty-aware reuse of earlier searches for query-efficient neural architecture search}
}
\newline
\\
Brama Seta Putra Satria\textsuperscript{1},
Dwi Sunaryono\textsuperscript{1*},
Misbakhul Munir Irfan Subakti\textsuperscript{1},
Irzal Ahmad Sabilla\textsuperscript{2},
Laras Suciningtyas\textsuperscript{3}
\\
\bigskip
\textbf{1} Department of Informatics, Institut Teknologi Sepuluh Nopember, Surabaya, East Java, Indonesia
\\
\textbf{2} Department of Information Technology, Institut Teknologi Sepuluh Nopember, Surabaya, East Java, Indonesia
\\
\textbf{3} Department of Electrical Engineering, Institut Teknologi Sepuluh Nopember, Surabaya, East Java, Indonesia
\\
\bigskip

* dwi@its.ac.id

\end{flushleft}

\section*{Abstract}
Neural architecture search (NAS) automates the design of neural networks, but on a new dataset it often affords only tens of short training runs, so what the search knows before its first evaluation decides its efficiency. Yet most NAS methods start from scratch, guided at best by zero-cost proxies (scores computed from an untrained network) that are unreliable across tasks. QuaNAS treats a library of earlier searches in the same search space as explicit knowledge of uncertain reliability. Main effects of design decisions learned from this library and, secondarily, zero-cost proxies become priors over candidate architectures without any label of the new dataset. A Gaussian process starts at the prior optimum (the candidate the priors rank highest) and re-estimates after every evaluation how far each prior is trusted. Under a protocol frozen before any confirmation data were seen, QuaNAS was evaluated once on 12 new image, time-series, text and tabular datasets. It lowered anytime regret (the normalized gap between the best architecture found so far and the best candidate, averaged over the search) on the short-training objective 2.7- to 3.9-fold relative to five pre-specified NAS baselines that start from scratch (Holm-adjusted $p=0.0012$ each). Its final models matched (non-inferiority margin 0.5 percentage points) or exceeded reference models trained with the same recipe on nine datasets, as did the prior optimum, selected without a single evaluation. Post-hoc comparisons with established transfer methods given the same library trace the gain to the library: without it, the same Gaussian process had over twice the regret. They also show where each transfer mechanism works best: the pooled main effects of QuaNAS for weakly transferable tabular tasks, per-task surrogates for image, time-series and text tasks and for small or mismatched libraries. Finally, the main effects make the transferred knowledge inspectable.

\clearpage
\newgeometry{top=0.85in,left=1in,right=1in,footskip=0.75in}

\section*{Introduction}\label{sec:intro}

Neural architecture search (NAS) automates the design of neural networks, but every candidate must be trained before its quality is known~\cite{elsken2019survey,white2023thousand,liu2023evonas}. Early NAS trained thousands of networks with reinforcement learning or evolution~\cite{zoph2017nas,real2019rea}, and weight sharing cut the cost at the price of biased estimates~\cite{liu2019darts}. When a practitioner needs a model for a new dataset, the realistic budget is a few dozen to a hundred short training runs. With such budgets, the order of evaluations matters more than asymptotic convergence, and simple strategies such as random search, local search and regularized evolution remain hard to beat~\cite{li2019random,yang2020frustrating,white2021ls,real2019rea}. Predictor-based Bayesian optimization (BO) such as BANANAS can improve on them~\cite{white2021bananas}. What a search knows \emph{before} it has trained anything therefore decides much of its efficiency.

Two kinds of cheap knowledge are available. The first, and in our results the more valuable, is earlier searches. Practitioners rarely search only once, and design choices that worked on one dataset often work on related ones. The evaluations of earlier searches in the same search space therefore form a library that a new search can reuse. Meta-learning for hyperparameter optimization exploits this by warm-starting BO with configurations that worked on similar datasets~\cite{feurer2015metalearning}, by weighting the surrogates of earlier tasks according to their ranking accuracy on the new one (ranking-weighted Gaussian process ensembles, RGPE)~\cite{feurer2018rgpe}, by sharing a learned basis or a pre-trained Gaussian process (GP) prior across tasks~\cite{perrone2018ablr,wang2024hyperbo}, or by prior-fitted networks that perform BO in context~\cite{muller2023pfns4bo}. Bai et al.~\cite{bai2023tlbo} survey this area. Transfer NAS methods meta-learn dataset encoders and architecture generators (MetaD2A)~\cite{lee2021metad2a}, deep-kernel GPs conditioned on a dataset encoding (TNAS)~\cite{shala2023tnas} or dataset-aware diffusion generators (DiffusionNAG)~\cite{an2024diffusionnag}, or map datasets to a knowledge base of architectures (KEGNAS)~\cite{wang2025kegnas}. Knowledge-assisted evolutionary algorithms have also been proposed for NAS~\cite{an2023krea}. BO over categorical spaces supplies the surrogate machinery, from overlap and Hamming kernels to graph kernels~\cite{oh2019combo,wan2021casmo,papenmeier2023bounce,ru2021nasbowl,garnett2023bo}.

\begin{figure}[!ht]
\centering
\includegraphics[width=0.68\linewidth]{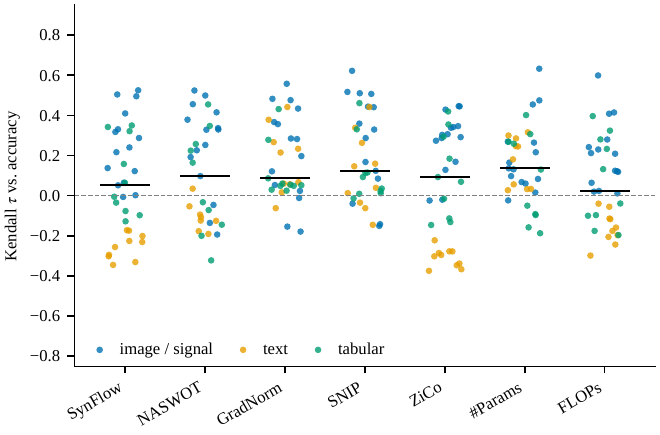}
\caption{\textbf{Zero-cost proxies are task-dependent.} Kendall rank correlation $\tau$ between each proxy and short-training accuracy on the 35 development pools (dots, colored by search space; black bars are medians). Every proxy is the best one on some pool, 19 pools contain a proxy with negative $\tau$, and on 15 pools even the best proxy has $\tau<0.3$. \#Params: parameter count; FLOPs: floating-point operations.}
\label{fig:motivation}
\end{figure}

The second kind is zero-cost proxies, which score an untrained network from a single mini-batch: SNIP and SynFlow through parameter saliencies~\cite{lee2019snip,tanaka2020synflow,abdelfattah2021zc}, NASWOT through activation codes~\cite{mellor2021naswot}, ZiCo through gradient statistics~\cite{li2023zico}, MeCo and SWAP through feature correlations and activation patterns~\cite{jiang2023meco,peng2024swap}, and ensembles or learned parametric proxies~\cite{lee2024aznas,dong2025parzc}. Large studies found that no proxy is reliable across search spaces and tasks, that parameter counts (\#Params in \cref{fig:motivation}) are often competitive~\cite{krishnakumar2022suitezero,li2024zeroshot}, and that training-free metrics are weaker still for text models~\cite{serianni2023rnn}. Our own development pools show the same picture (\cref{fig:motivation}). Proxy-calibrating searches such as ProxyBO, HNAS and RoBoT therefore weight proxies by their agreement with the evaluations of the current search~\cite{shen2023proxybo,shu2022hnas,he2024robot}, but every search starts from scratch. Proxies are thus best treated as a secondary source, to fall back on when no informative earlier search exists.

Tabular and surrogate benchmarks made NAS methods comparable by lookup~\cite{dong2020nb201,duan2021transnas,krishnakumar2022suitezero}, but they also made it easy to tune and report on the same tasks~\cite{lindauer2020best,yang2020frustrating}. NAS-Bench-360 also showed that methods developed on vision rarely beat expert architectures on other modalities~\cite{tu2022nb360}. Recent generators that propose architectures with large language models or diffusion models~\cite{chen2023evoprompting,nasir2024llmatic,an2024diffusionnag} inherit the same evaluation questions.

\subsection*{Gap analysis and motivation}\label{sec:intro:gaps}

Four gaps in the existing literature jointly motivate the present work.
\begin{enumerate}[label=\arabic*.,leftmargin=*]
\item \textbf{Reused knowledge is neither checked nor inspectable.} Transfer NAS methods reuse earlier tasks through meta-trained encoders whose transferred knowledge cannot be read off, and ranking-weighted ensembles from hyperparameter optimization judge earlier tasks only by their agreement with the evaluations of the new task~\cite{feurer2018rgpe}. To our knowledge, no NAS method (i) checks before the first evaluation whether a small library of earlier searches is informative at all, (ii) states which design choices it transfers, (iii) re-estimates its trust in them from every evaluation, and (iv) falls back to other cheap knowledge, such as zero-cost proxies, when the library is not informative.
\item \textbf{Transfer needs machinery that small libraries cannot feed.} Transfer NAS methods rely on dataset encoders or knowledge bases meta-trained on many tasks in the same search space. A practitioner with a handful of earlier searches has no such collection, and the simpler transfer methods of hyperparameter optimization have rarely been evaluated for NAS.
\item \textbf{Evaluation on the benchmarks used for design.} NAS methods are often tuned and evaluated on the same benchmarks. The seeds of nearly deterministic searches are treated as independent replicates, which inflates significance, and claims are rarely confirmed on held-out datasets under criteria fixed in advance~\cite{lindauer2020best,yang2020frustrating,nosek2018prereg,kapoor2023leakage}.
\item \textbf{Vision-centric evidence.} Query-efficient NAS is evaluated mostly on vision benchmarks~\cite{tu2022nb360}. Whether its gains hold on text, tabular and time-series data, and on architectures never seen during design, is largely untested.
\end{enumerate}
These gaps lead to our research question: \emph{can a search reuse a small library of earlier searches without labels of the new task, check and re-estimate how far this knowledge can be trusted, show what it transfers, and keep its advantage on datasets and architectures that played no role in its design?} \Cref{tab:positioning} positions the present work against representative methods. Among them, only the method proposed here combines both kinds of cheap knowledge and re-weights them from the evaluations of the new task.

\begin{table}[!ht]
\centering
\caption{\textbf{Positioning of QuaNAS against representative prior work, based on the original articles.}}
\label{tab:positioning}
\footnotesize
\setlength{\tabcolsep}{4pt}
\begin{tabularx}{\textwidth}{@{}>{\raggedright\arraybackslash}p{4.2cm}>{\centering\arraybackslash}p{1.25cm}>{\centering\arraybackslash}p{1.25cm}>{\centering\arraybackslash}p{1.6cm}>{\raggedright\arraybackslash}X>{\raggedright\arraybackslash}X@{}}
\toprule
Method & Proxies & Earlier tasks & Online re-weighting & Task representation needed & Data in the original work \\
\midrule
REA~\cite{real2019rea}, LS~\cite{white2021ls} & $\times$ & $\times$ & -- & none & image benchmarks \\
BANANAS~\cite{white2021bananas} & $\times$ & $\times$ & -- & none & image benchmarks \\
ProxyBO~\cite{shen2023proxybo} & \checkmark & $\times$ & \checkmark & none & image benchmarks \\
RoBoT~\cite{he2024robot} & \checkmark & $\times$ & \checkmark & none & image benchmarks \\
Warm-start BO~\cite{feurer2015metalearning} (HPO) & $\times$ & \checkmark & $\times$ & dataset meta-features & hyperparameter benchmarks \\
RGPE~\cite{feurer2018rgpe} (HPO) & $\times$ & \checkmark & \checkmark & none & hyperparameter benchmarks \\
HyperBO~\cite{wang2024hyperbo} (HPO) & $\times$ & \checkmark & $\sim$ & none & hyperparameter benchmarks \\
MetaD2A~\cite{lee2021metad2a} & $\times$ & \checkmark & $\times$ & meta-trained dataset encoder & image datasets \\
TNAS~\cite{shala2023tnas} & $\times$ & \checkmark & \checkmark & meta-trained dataset encoder & image datasets \\
KEGNAS~\cite{wang2025kegnas} & $\times$ & \checkmark & $\sim$ & dataset encoding & graph datasets \\
\textbf{QuaNAS (ours)} & \checkmark & \checkmark & \checkmark & modality detected from raw files & image, time series, text, tabular \\
\bottomrule
\end{tabularx}
\begin{flushleft}\footnotesize \checkmark: yes; $\times$: no; $\sim$: partly (the posterior is updated, but the transferred prior itself is not re-weighted). ``Online re-weighting'' means that the weight of the cheap knowledge (proxies or earlier tasks) is adapted from evaluations on the new task. HPO: proposed for hyperparameter optimization. REA: regularized evolution; LS: local search.
\end{flushleft}
\end{table}

\subsection*{Contributions}\label{sec:intro:contrib}

The method proposed here, QuaNAS (\textbf{Q}uery-efficient, \textbf{u}ncertainty-\textbf{a}ware NAS), and its evaluation contribute the following.
\begin{enumerate}[label=\arabic*.,leftmargin=*]
\item \textbf{Target-label-free transfer from earlier searches.} Earlier searches are reused through ridge main-effect models of a shared architecture encoding, pooled over the library and weighed against zero-cost proxies by leave-one-task-out stacking, without labels of the new task, dataset encoders or domain labels. The resulting prior alone, without any evaluation, selected architectures that matched (within a non-inferiority margin of 0.5 percentage points [pp]) or exceeded the matched-recipe reference models on 9 of 12 held-out datasets. Removing it raised regret significantly, and its main effects show which design choices carried over and how consistently (see ``\nameref{sec:res:ablation}'' and ``\nameref{sec:res:transfer}'').
\item \textbf{A Gaussian-process search seeded and re-weighted by uncertain priors.} The search starts at the prior optimum, the prior weights enter the GP as random effects that every evaluation updates, and a reliability switch discards a transferred prior that the library itself shows to be uninformative. The prior-driven start and the Hamming kernel were necessary on held-out data, whereas re-weighting and the switch acted as safeguards against large failures during development (see ``\nameref{sec:method:search}'' and ``\nameref{sec:res:ablation}'').
\item \textbf{Evidence on how earlier searches should be reused.} Under a frozen protocol and on 12 new datasets from four modalities, QuaNAS lowered anytime regret 2.7- to 3.9-fold relative to five pre-specified baselines that start from scratch (Holm-adjusted $p=0.0012$ against each), and its final models matched or exceeded matched-recipe reference models on 9 of the 12 datasets. Post-hoc comparisons with established transfer methods given the same library trace this gain to earlier searches rather than proxies. They also show where each way of reusing them works best: pooled main effects, as in QuaNAS, for weakly transferable tabular tasks and per-task surrogates for image, time-series and text tasks and for small or mismatched libraries (see ``\nameref{sec:res:search}'', ``\nameref{sec:res:transfer-bo}'', ``\nameref{sec:res:robust}'' and ``\nameref{sec:res:models}'').
\item \textbf{An open resource} of 47 candidate pools with real short-training accuracies, per-proxy costs and full-training results, together with all code and per-run results (see the Data Availability statement).
\end{enumerate}

The main takeaway is that, at budgets of about one hundred short trainings, earlier searches rather than zero-cost proxies are the knowledge that matters, and that main effects make this knowledge inspectable. The remainder of this paper is organized as follows. Materials and methods presents QuaNAS and the experimental protocol, Results reports the results on the held-out datasets, including post-hoc comparisons with transfer methods and a standard benchmark, Discussion discusses them, and Conclusion concludes.

\section*{Materials and methods}\label{sec:method}

\subsection*{Problem formulation}\label{sec:problem}

A search space $\mathcal{X}=\prod_{d=1}^{D}\{1,\dots,n_d\}$ encodes an architecture $x$ as $D$ categorical decisions, such as the depth, width and block type of each stage; $\phi(x)\in\{0,1\}^{P}$, $P=\sum_d n_d$, is its one-hot encoding. A \emph{task} $T$ is a dataset together with a search space and a candidate pool $\pool=\{x_1,\dots,x_N\}\subseteq\mathcal{X}$, a sample of $\mathcal{X}$ (uniform, or stratified by one decision) or the whole space when it is small; the task is the unit of analysis. Querying $x$ trains it with a fixed short recipe and returns its validation accuracy $y(x)$, the \emph{low-fidelity} accuracy, at a measured cost of $c(x)$ seconds. A search may query at most $B$ distinct pool members $x_{(1)},x_{(2)},\dots$. It may also compute zero-cost proxies $p_j(x)$, $j=1,\dots,J$, at cost $c_j(x)$ without training, and it may use a \emph{library} $\lib$ of earlier tasks $u$ in the same search space, each with its pool $\pool_u$, low-fidelity accuracies $y_u$ and proxy scores. Labels of the new task are available only through its own queries. We assume that encodings are shared by all tasks of a search space and that low-fidelity accuracy ranks architectures consistently across training seeds, which we verify per task (see ``\nameref{sec:protocol:pools}''). \Cref{tab:notation} lists the notation.

\begin{table}[!ht]
\centering
\caption{\textbf{Main notation.}}
\label{tab:notation}
\footnotesize
\setlength{\tabcolsep}{3pt}
\begin{tabularx}{\columnwidth}{@{}l>{\raggedright\arraybackslash}X@{}}
\toprule
Symbol & Meaning \\
\midrule
$x$, $D$, $n_d$ & architecture, number of decisions, options of decision $d$ \\
$\pool$, $N$, $B$ & candidate pool of the new task, its size, query budget (100) \\
$y(x)$, $c(x)$ & low-fidelity validation accuracy and its training cost \\
$\lib$, $u$ & library of earlier tasks in the same space, one library task \\
$\ns(\cdot)$ & rank-normal transform (ranks mapped to normal quantiles) \\
$a(x)$, $s(x)$ & architecture prior (from $\lib$) and proxy prior \\
$\mu_a$, $\mu_s$ & prior weights meta-learned on $\lib$ \\
$\beta_a$, $\beta_s$, $\beta_0$ & random prior coefficients and offset in the GP \\
$\sigma_a$, $\sigma_s$, $\sigma_c$ & prior standard deviations of $\beta_a$, $\beta_s$, $\beta_0$ \\
$\sigma_{\mathrm{H}}$, $\theta$, $\sigma_n$ & scale and rate of the Hamming kernel, noise standard deviation \\
$\rho$ & fraction of the pool scored by proxies (0.2) \\
$\nreg_q$, $\AUC$, $\TTT$ & regret after $q$ queries, anytime regret, time to target \\
\bottomrule
\end{tabularx}
\end{table}

Search quality after $q$ queries is the normalized regret
\begin{equation}
\nreg_q=\frac{y^\star-\max_{t\le q}y(x_{(t)})}{y^\star-\tilde y},
\label{eq:nreg}
\end{equation}
where $y^\star$ and $\tilde y$ are the maximum and the median of $y$ over $\pool$ (the minimum replaces the median if the median equals the maximum). It is 0 once the best pool member has been found and about 1 when the best query is only as good as the median candidate. When the pool is a uniform sample, its best member lies in expectation in the top $1/(N+1)$ of the space (the top 0.17\% for $N=600$), so zero regret means finding a top architecture of the space, not merely the best of a small set. The primary endpoint is the anytime regret, the area under the regret curve,
\begin{equation}
\AUC=\frac{1}{B}\sum_{q=1}^{B}\nreg_q ,
\label{eq:auc}
\end{equation}
which rewards finding good architectures early. Search speed is the time to target
\begin{equation}
\TTT=\sum_{t\le q^\star}c\big(x_{(t)}\big)+\sum_{x\in S_{q^\star}}\sum_{j\in\mathcal{J}}c_j(x)+o_{q^\star},
\label{eq:ttt}
\end{equation}
where $q^\star=\min\{q:\nreg_q\le0.05\}$ is the first query that comes within 5\% of the best--median gap, $S_{q^\star}$ is the set of architectures whose proxies $\mathcal{J}$ the method has computed before query $q^\star$ and $o_{q^\star}$ is the measured algorithmic overhead. Runs that never reach the target are censored at 1.01 times the largest total cost of any run on that task. Both endpoints are defined on the low-fidelity accuracy that the search observes; the architecture finally reported is trained with a longer \emph{full} recipe and evaluated once on the test split, and is assessed separately (see ``\nameref{sec:res:models}'').

\subsection*{Design overview: QuaNAS}\label{sec:method:overview}

\begin{figure}[!ht]
\centering
\includegraphics[width=\linewidth]{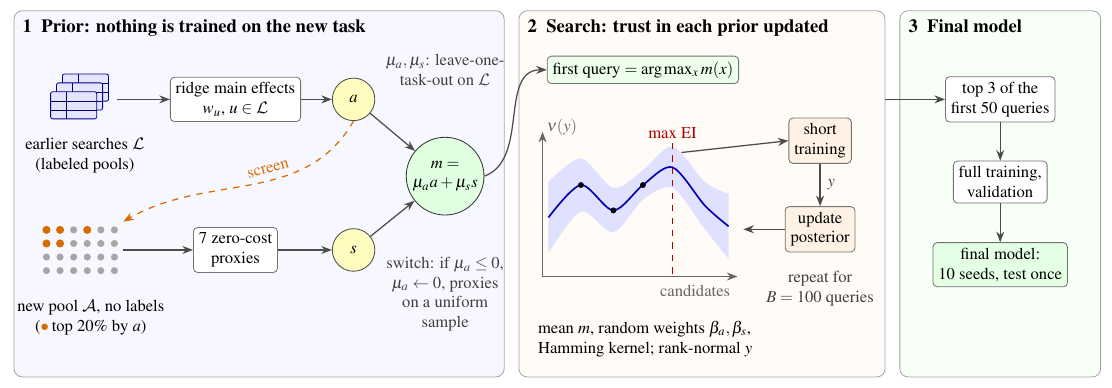}
\caption{\textbf{Overview of QuaNAS.} (1) The architecture prior $a$ comes from earlier searches in the same space, and the proxy prior $s$ from zero-cost proxies computed only for the fifth of the pool that $a$ ranks highest; their weights are meta-learned on the library, and nothing is trained on the new task. (2) The search starts at the prior optimum and continues with expected improvement (EI); after every short training, the posterior, and with it the weight of each prior, is updated. (3) A small full-training shortlist selects the reported model.}
\label{fig:overview}
\end{figure}

QuaNAS (\textbf{Q}uery-efficient, \textbf{u}ncertainty-\textbf{a}ware NAS; \cref{fig:overview} and \cref{alg:quanas}) proceeds in three stages. It first builds two priors over the pool without training anything on the new task (see ``\nameref{sec:method:prior}''), then spends the budget with a Gaussian process (GP) whose prior mean is the weighted combination of the two priors and whose random coefficients let the new evaluations re-weight that combination (see ``\nameref{sec:method:search}''), and finally selects the reported model with a small full-training shortlist (see ``\nameref{sec:method:final}''). Unlike ranking-weighted ensembles, which weight per-task surrogates only by their fit to the new task's evaluations~\cite{feurer2018rgpe}, QuaNAS weights its transferred prior before the first evaluation, by how well the library predicts its own held-out members, and then lets every evaluation re-weight it.

\begin{algorithm}[!ht]
\caption{QuaNAS}\label{alg:quanas}
\begin{algorithmic}[1]
\Require raw files and pool $\pool$ of task $T$; library tables $(\pool_u,y_u,p_u)$; budget $B$
\Ensure query sequence; final model and its test accuracy
\State $\ell\gets$ \Call{DetectModality}{raw files}; select space, recipe and $\lib=\{u:\mathrm{space}_u=\mathrm{space}_T\}$
\State fit $w_u$, $u\in\lib$ (\cref{eq:ridge}); $a\gets\ns(\frac{1}{|\lib|}\sum_u\boldsymbol{X}w_u)$ \Comment{\cref{eq:aprior}}
\State $S\gets$ top $\lceil0.2N\rceil$ of $\pool$ by $a$; proxies on $S$; $s$ by \cref{eq:sprior}
\State $(\mu_a,\mu_s)\gets$ leave-one-task-out stacking on $\lib$ \Comment{\cref{eq:meta}}
\If{$\mu_a\le0$} apply the reliability switch \Comment{\cref{eq:switch}} \EndIf
\State $m\gets\mu_a a+\mu_s s$; query $x_{(1)}=\arg\max_{x}m(x)$; $Q\gets\{x_{(1)}\}$
\For{$t=1,\dots,B-1$}
  \State $z_Q\gets\ns(y_Q)$; $\mu_t$, $\sigma_t$ on $\pool\setminus Q$ by \cref{eq:posterior}
  \State short-train $x_{(t+1)}=\arg\max_{x\in\pool\setminus Q}\EI_t(x)$; add it to $Q$
\EndFor
\State $C\gets$ top 3 of the first 50 queries by $y$; full-train each; $x^\dagger\gets$ best on validation
\State full-train $x^\dagger$ with seeds 1--10; open the test split once
\end{algorithmic}
\end{algorithm}

\subsection*{Target-label-free priors from earlier searches and zero-cost proxies}\label{sec:method:prior}

\begin{definition}[Target-label-free prior]\label{def:labelfree}
A prior over the pool $\pool$ of a task is \emph{target-label-free} if it is a function of the encodings of $\pool$, of proxy scores on $\pool$, of the detected modality and of the library tables $\{(\pool_u,y_u,p_u)\}_{u\in\lib}$, but not of any accuracy of the task itself.
\end{definition}

All quantities in this subsection are target-label-free: the implementation builds them from a record of the new task that contains no accuracies, and a unit test verifies that changing those accuracies leaves them unchanged.

\textbf{Modality detection and library.}
QuaNAS detects the data modality from an unlabeled sample of the raw training files with deterministic rules (image files, JSON-lines or tables with long string fields, arrays with two to four axes). The modality selects the search space (convolutional for images and signals, a text space, a tabular space) and its training recipe, and the library $\lib$ contains all earlier tasks of that search space. The user supplies no domain label.

\textbf{Architecture prior.}
Let $\boldsymbol{X}_u$ and $\boldsymbol{X}$ stack the encodings $\phi(x)$ of $\pool_u$ and $\pool$. For each library task we fit a ridge main-effect model to rank-normal accuracies,
\begin{equation}
w_u={\big(\boldsymbol{X}_u^{\top}\boldsymbol{X}_u+\lambda I\big)}^{-1}\boldsymbol{X}_u^{\top}\ns(y_u),\qquad \lambda=1,
\label{eq:ridge}
\end{equation}
where $\ns(v)_i=\Phi^{-1}\!\big((r_i-\tfrac12)/n\big)$ maps the average ranks $r_i$ of a vector $v$ of length $n$ to standard-normal quantiles. The model assigns an additive effect to every option of every decision, and the rank-normal transform makes tasks with different accuracy ranges comparable. The architecture prior of the new pool is the rank-normalized average prediction
\begin{equation}
a=\ns\Big(\tfrac{1}{|\lib|}\textstyle\sum_{u\in\lib}\boldsymbol{X}w_u\Big).
\label{eq:aprior}
\end{equation}
Main effects have few parameters ($P\le41$), so one earlier pool suffices to estimate them; they apply to architectures that no library task has evaluated and state the transferred knowledge explicitly (see ``\nameref{sec:res:transfer}''). The average weights all library tasks equally, without a task-similarity measure; whether the library is informative at all is checked by the stacking below.

\textbf{Screened proxy prior.}
Proxies are cheap per architecture but not free, so QuaNAS computes them only for the screened set $S$ of the $\lceil\rho N\rceil$ candidates with the largest $a$ ($\rho=0.2$). For every active proxy $j$ (a proxy with more than 20\% non-finite values on the pool is dropped), $e_{ij}$ is the empirical distribution function of $p_j(x_i)$ within $S$, with average ranks and $e_{ij}=0.5$ for non-finite values, and
\begin{equation}
s_i=\begin{cases}\ns\big(\textstyle\sum_j e_{\cdot j}\big)_i, & i\in S,\\ 0, & i\notin S .\end{cases}
\label{eq:sprior}
\end{equation}

\textbf{Meta-learned prior weights.}
The weights of the two priors estimate how well each predicts a held-out task of the library. For every $u\in\lib$, $a_u$ and $s_u$ are recomputed on $\pool_u$ from the library $\lib\setminus\{u\}$, and
\begin{equation}
(\mu_a,\mu_s)=\arg\min_{\mu_a,\mu_s}\sum_{u\in\lib}{\big\lVert\ns(y_u)-\mu_a a_u-\mu_s s_u\big\rVert}_2^2 .
\label{eq:meta}
\end{equation}

\textbf{Reliability switch.}
A non-positive $\mu_a$ means that, across the library, the transferred architecture prior does not predict held-out tasks. It should then neither steer the start nor decide which candidates receive proxy scores, and QuaNAS sets
\begin{equation}
\mu_a\leftarrow0,\qquad S\leftarrow U_{\min(N,2000)}(\pool),\qquad \mu_s\leftarrow\frac{\sum_{u}\bar s_u^{\top}\ns(y_u)}{\sum_{u}{\lVert\bar s_u\rVert}_2^{2}},
\label{eq:switch}
\end{equation}
where $U_m(\pool)$ is a uniform sample of $m$ candidates (the whole pool if $N\le2000$, drawn with the same random stream as the proxy-ranking baseline), $s$ is recomputed on the new $S$ by \cref{eq:sprior}, and $\bar s_u$ is the proxy prior of library task $u$ computed on its whole pool. The architecture prior stays in the GP kernel, where the new task's evaluations can still give it weight.

\subsection*{Search with uncertain prior weights}\label{sec:method:search}

\textbf{Gaussian process with random prior weights.}
QuaNAS models the rank-normal accuracy of the new task as
\begin{equation}
\begin{gathered}
f(x)=\beta_a a(x)+\beta_s s(x)+\beta_0+g(x),\\
\beta_a\sim\mathcal{N}(\mu_a,\sigma_a^2),\quad \beta_s\sim\mathcal{N}(\mu_s,\sigma_s^2),\quad \beta_0\sim\mathcal{N}(0,\sigma_c^2),
\end{gathered}
\label{eq:model}
\end{equation}
where $g$ is a zero-mean GP with the exponential Hamming kernel $\sigma_{\mathrm{H}}^2\exp\!\big(-\theta H(x,x')\big)$ and $H$ is the number of differing decisions. The Hamming kernel captures local structure and interactions that the additive priors miss. Integrating out the coefficients gives a GP with mean and covariance
\begin{equation}
\begin{aligned}
m(x)&=\mu_a a(x)+\mu_s s(x),\\
k(x,x')&=\sigma_a^2a(x)a(x')+\sigma_s^2s(x)s(x')+\sigma_c^2+\sigma_{\mathrm{H}}^2e^{-\theta H(x,x')}.
\end{aligned}
\label{eq:kernel}
\end{equation}
After $t$ queries with index set $Q$, the observations $z_Q=\ns(y_Q)$ are recomputed, treated as $f(x_Q)$ plus Gaussian noise of variance $\sigma_n^2$, and give the posterior
\begin{equation}
\begin{aligned}
\mu_t(x)&=m(x)+k_{xQ}{(K_{QQ}+\sigma_n^2I)}^{-1}(z_Q-m_Q),\\
\sigma_t^2(x)&=k(x,x)-k_{xQ}{(K_{QQ}+\sigma_n^2I)}^{-1}k_{Qx}.
\end{aligned}
\label{eq:posterior}
\end{equation}

\textbf{Two properties.}
The random coefficients make the trust in each prior a quantity that the new evaluations update; fixing $\sigma_a=\sigma_s=0$ instead keeps the meta-learned weights for the whole search. \Cref{rem:calib} makes this explicit, and \cref{rem:invariance} states why tasks with different accuracy ranges can be pooled.

\begin{remark}[Precision-weighted recalibration]\label{rem:calib}
With $h(x)={(a(x),s(x),1)}^{\top}$, $\boldsymbol{R}_Q=[h(x)]_{x\in Q}$ and $K_g$ the covariance of $g$ on $Q$ plus $\sigma_n^2I$, \cref{eq:model} is a GP with explicit basis functions~\cite[Sec.~2.7]{rasmussen2006gp}. The posterior of $\beta=(\beta_a,\beta_s,\beta_0)$ has precision $\Lambda=\Sigma_\beta^{-1}+M$, $M=\boldsymbol{R}_QK_g^{-1}\boldsymbol{R}_Q^{\top}$, $\Sigma_\beta=\operatorname{diag}(\sigma_a^2,\sigma_s^2,\sigma_c^2)$, and mean $\bar\beta=\Lambda^{-1}(\Sigma_\beta^{-1}b+M\hat\beta)$, where $b={(\mu_a,\mu_s,0)}^{\top}$ and $\hat\beta$ is the generalized least-squares fit to the new task's evaluations alone. Hence $\bar\beta=b$ before any evaluation and $\bar\beta$ moves toward $\hat\beta$ as evaluations accumulate, at a rate set by the prior precisions $1/\sigma_a^2$ and $1/\sigma_s^2$.
\end{remark}

\begin{remark}[Invariance to accuracy scales]\label{rem:invariance}
Every input of QuaNAS depends on accuracies only through average ranks ($\ns$ in \cref{eq:ridge,eq:meta,eq:switch,eq:posterior}). Strictly increasing transformations of the accuracies of any library task or of the new task therefore leave the priors, their weights, the switch decision, every query and the final shortlist unchanged.
\end{remark}

A prior that misleads on the new task therefore loses weight, and its weight can even change sign. With $\sigma_a=\sigma_s=1$ the meta-learned weights are only weakly anchored, which matters because they can themselves be wrong for an unusual task. Earlier tasks with very different accuracy ranges, for instance 60\% and 97\%, are therefore pooled without any normalization choice.

\textbf{Prior-driven start and acquisition.}
The first query is the prior optimum $x_{(1)}=\arg\max_{x\in\pool}m(x)$; there is no random initial design. Every later query maximizes expected improvement over all unqueried candidates~\cite{jones1998ego},
\begin{equation}
\EI_t(x)=\big(\mu_t(x)-z^{+}\big)\Phi(\gamma)+\sigma_t(x)\varphi(\gamma),\qquad \gamma=\frac{\mu_t(x)-z^{+}}{\sigma_t(x)},
\label{eq:ei}
\end{equation}
where $z^{+}=\max z_Q$ and $\Phi$, $\varphi$ are the standard-normal distribution and density functions. Exact ties are broken by a jitter of order $10^{-9}$ from the run's random stream, so for a given task the search is deterministic up to tie-breaking.

\textbf{Hyperparameters.}
All tasks use $\sigma_a=\sigma_s=1$, $\sigma_c^2=\sigma_{\mathrm{H}}^2=0.25$, $\theta=0.5$, $\sigma_n^2=0.1$, $\lambda=1$ and $\rho=0.2$. They were fixed on the development pools and are not refitted by marginal likelihood: with at most 100 rank-normal observations, fixed values avoid unstable fits and keep the search deterministic. Marginal-likelihood fitting, Bayesian averaging of the hyperparameters, upper-confidence-bound acquisition and acquisition portfolios were tried during development and did not lower development regret; the ``\nameref{sec:res:ablation}'' section reports the sensitivity to the fixed values.

\subsection*{Selection of the final model}\label{sec:method:final}

Short-training accuracies of the best candidates often differ by less than their noise. QuaNAS therefore shortlists the three candidates with the highest low-fidelity accuracy among its first 50 queries, the same half of the budget from which the model-level comparators take their architectures (see ``\nameref{sec:protocol:methods}''), trains each once with the full recipe and keeps the one with the highest \emph{validation} accuracy (ties: higher low-fidelity accuracy, then earlier query). The selected architecture is trained with ten further seeds and evaluated on the test split; the selection seed is not reused. The search and its regret are unaffected by this step.

\subsection*{Computational complexity}\label{sec:method:complexity}

Fitting \cref{eq:ridge} costs $O(N_uP^2+P^3)$ per library task, and the leave-one-task-out stacking of \cref{eq:meta} repeats the prior construction $|\lib|$ times; both are done once per space. Proxies are computed for $\lceil\rho N\rceil$ candidates instead of all $N$, or 2000, as in proxy ranking. With the Hamming matrix precomputed in $O(N^2D)$, the $t$-th query costs $O(t^3+Nt^2)$, i.e.\ $O(B^4+NB^3)$ per search, which is negligible for $B=100$ and $N\le3456$. Measured on one CPU thread, the whole algorithmic overhead of a QuaNAS run averaged 0.28\,s (maximum 0.95\,s), whereas a single short training took 6.0\,s on average on the GPU (see ``\nameref{sec:protocol:training}'').

\subsection*{Experimental protocol}\label{sec:protocol}

\subsubsection*{Development and confirmation datasets}\label{sec:protocol:data}

\textbf{Development pools.} All design decisions were made on 35 labeled pools from earlier development cycles of this project: ten convolutional pools (EuroSAT~\cite{helber2019eurosat}, Malaria~\cite{rajaraman2018malaria}, Spherical CIFAR-100, Satellite and NinaPro~\cite{tu2022nb360}, CIFAR-10~\cite{krizhevsky2009cifar}, Fashion-MNIST~\cite{xiao2017fmnist}, SVHN~\cite{netzer2011svhn}, BloodMNIST~\cite{yang2023medmnist}, UCI-HAR~\cite{anguita2013har}), four text and four tabular pools in an earlier version of the text and tabular spaces (DBPedia and AG News~\cite{zhang2015charcnn}, IMDB~\cite{maas2011imdb}, TREC~\cite{li2002trec}; Dry Bean~\cite{koklu2020drybean}, Letter~\cite{slate1991letter}, Adult~\cite{becker1996adult}, Covertype~\cite{blackard1998covertype}), the same eight datasets in the final spaces, and nine further pools (KMNIST~\cite{clanuwat2018kmnist}, OrganAMNIST and PathMNIST~\cite{yang2023medmnist}, GTSRB~\cite{stallkamp2012gtsrb}, FordA~\cite{dau2019ucr}, Emotion~\cite{saravia2018carer}, Yelp Polarity~\cite{zhang2015charcnn}, MAGIC~\cite{bock2004magic}, Connect-4~\cite{tromp1995connect4}). The pools in the final spaces also form the library from which QuaNAS transfers knowledge to the confirmation tasks (15 convolutional, 6 text and 6 tabular pools).

\textbf{Confirmation datasets.} The frozen method was then run once on 12 datasets that had never been loaded in any earlier cycle (\cref{tab:datasets}): four image, two time-series, three text and three tabular classification tasks. They were chosen as public benchmarks with official or published splits, full training of a candidate in under about five minutes on one GPU, and the same modality mix as earlier cycles; ease of beating the baselines played no role. The datasets are new to the project, but several come from collections that also supplied development pools (MedMNIST, the UCR archive, the text collection of Zhang et al.~\cite{zhang2015charcnn}); the ``\nameref{sec:res:robust}'' section tests whether the transfer depends on such related datasets. Three of the originally selected datasets were replaced by fallback datasets before any search was run: Banking77 could not be downloaded and its first fallback (tweet\_eval sentiment) failed the pool quality check, Crop failed it, and DermaMNIST and its first fallback (PneumoniaMNIST) failed it, each after the specified repair; Amazon Polarity, FetalECG (UCR NonInvasiveFetalECGThorax1) and OCTMNIST took their places. All replacements followed from download or quality-control outcomes alone, without any accuracy of these datasets being seen.

\begin{table}[!ht]
\centering
\caption{\textbf{Confirmation datasets.}}
\label{tab:datasets}
\footnotesize
\setlength{\tabcolsep}{3.0pt}
\begin{tabular}{@{}llrlrccc@{}}
\toprule
Dataset & Input & Classes & Train / val.\ / test & Pool & Short & Full & Stability \\
\midrule
\multicolumn{8}{@{}l}{\emph{Image} (convolutional space, library of 15 earlier pools)}\\
OCTMNIST\textsuperscript{\dag}~\cite{yang2023medmnist} & $28{\times}28{\times}1$ & 4 & 97,477 / 10,832 / 1,000 & 600 & 3 (20k) & 15 & 0.81\\
TissueMNIST~\cite{yang2023medmnist} & $28{\times}28{\times}1$ & 8 & 165,466 / 23,640 / 47,280 & 600 & 3 (20k) & 10 & 0.89\\
EMNIST-Balanced~\cite{cohen2017emnist} & $28{\times}28{\times}1$ & 47 & 101,520 / 11,280 / 18,800 & 600 & 3 (20k) & 15 & 0.96\\
Imagenette-32~\cite{howard2019imagenette} & $32{\times}32{\times}3$ & 10 & 8,523 / 946 / 3,925 & 600 & 5 & 60 & 0.65\\
\multicolumn{8}{@{}l}{\emph{Time series} (convolutional space, library of 15 earlier pools)}\\
ElectricDevices~\cite{dau2019ucr} & $1{\times}96$ & 7 & 7,587 / 1,339 / 7,711 & 600 & 10 & 60 & 0.72\\
FetalECG\textsuperscript{\dag}~\cite{dau2019ucr} & $1{\times}750$ & 42 & 1,531 / 269 / 1,965 & 600 & 10 & 60 & 0.92\\
\multicolumn{8}{@{}l}{\emph{Text} (text space, library of 6 earlier pools)}\\
Yahoo Answers~\cite{zhang2015charcnn} & $\le256$ tokens & 10 & 45,000 / 5,000 / 60,000 & 600 & 2 & 5 & 0.90\\
Amazon Polarity\textsuperscript{\dag}~\cite{zhang2015charcnn} & $\le256$ tokens & 2 & 45,000 / 5,000 / 400,000 & 600 & 2 & 5 & 0.89\\
20 Newsgroups~\cite{lang1995newsweeder} & $\le256$ tokens & 20 & 10,185 / 1,129 / 7,532 & 600 & 5 & 20 & 0.94\\
\multicolumn{8}{@{}l}{\emph{Tabular} (tabular space, library of 6 earlier pools)}\\
Chess KRK~\cite{bain1994krk} & 3 cat. + 3 num. & 18 & 19,637 / 15\% / 15\% & 3,456 & 15 & $\le100$ & 0.98\\
Grid Stability~\cite{arzamasov2018grid} & 12 num. & 2 & 7,000 / 1,500 / 1,500 & 3,456 & 15 & $\le100$ & 0.65\\
Credit Default~\cite{yeh2009credit} & 23 num. & 2 & 21,000 / 4,500 / 4,500 & 3,456 & 30 & $\le100$ & 0.57\\
\bottomrule
\end{tabular}
\begin{flushleft}\footnotesize Pool: number of candidate architectures (convolutional and text: sampled, disjoint from all development pools except the 18 bag-of-embeddings text architectures; tabular: the whole space). Short/full: training epochs of a query and of a final model (``20k'': fixed stratified subset of 20,000 training samples; tabular full training uses early stopping). Stability: Spearman correlation of short-training accuracy between two training seeds on 30 random pool members (quality-control threshold 0.5). cat./num.: categorical/numerical features. \textsuperscript{\dag}Fallback dataset (see ``\nameref{sec:protocol:data}'').
\end{flushleft}
\end{table}

\subsubsection*{Search spaces, pools and quality control}\label{sec:protocol:pools}

The convolutional space (2,834,352 architectures, built with 2-D or 1-D convolutions according to the modality) has 15 decisions: stem width $\{16,24,32\}$; for each of three stages the depth $\{1,2,3\}$, the width (three stage-specific options), the block type (MBConv with expansion 3 or 6, residual basic block) and the kernel size $\{3,5\}$; squeeze-and-excitation; and the activation (ReLU, SiLU, Hardswish). The text space (23,328 canonical architectures) has ten decisions: embedding dimension, unigram or unigram plus hashed-bigram input, encoder (bag of embeddings, TextCNN, residual 1-D convolution or BiGRU), layers, kernel, width, pooling, a bag skip connection, dropout and activation. The tabular space (3,456 architectures) has seven decisions: depth, width, residual connections, normalization, activation, dropout and periodic numerical embeddings~\cite{gorishniy2022embeddings}.

Each confirmation pool of the convolutional space is a uniform sample of 600 architectures, each text pool contains all 18 bag-of-embeddings architectures and 194 uniformly sampled architectures of each other encoder, and each tabular pool is the whole space. Apart from the bag-of-embeddings architectures, the convolutional and text pools of the confirmation tasks are disjoint from every development pool, so QuaNAS had to transfer to architectures it had never seen. For every pool member we computed seven zero-cost proxies at random initialization (SynFlow, NASWOT, GradNorm, SNIP, ZiCo, number of parameters and FLOPs) on one training mini-batch of 64 samples, together with the time of each, the parameter count, multiply--accumulate operations and serialized size, and the low-fidelity validation accuracy with its measured training time.

Each pool passed a method-independent quality check before any search: the low-fidelity accuracies had to spread by more than 0.5\,pp, and 30 random members retrained with a second seed had to reproduce the ranking with a Spearman correlation $r_s\ge0.5$. A failing pool was rebuilt once with twice the short-training epochs and replaced by its specified fallback if it failed again. During this check only pass or fail and the stability coefficient were reported; no accuracy of a confirmation task was seen before the protocol was frozen.

\subsubsection*{Leakage prevention and frozen protocol}\label{sec:protocol:prereg}

\begin{figure}[!ht]
\centering
\includegraphics[width=0.68\linewidth]{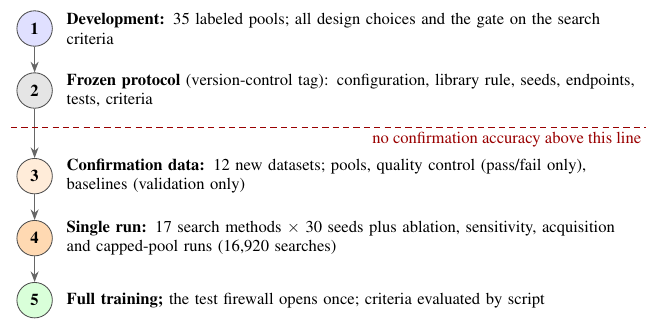}
\caption{\textbf{Development--confirmation protocol.} No accuracy of a confirmation task was available before the protocol was frozen, and every test split was read once, after all selections.}
\label{fig:protocol}
\end{figure}

\Cref{fig:protocol} shows the protocol. Official splits were used where available and stratified splits with a fixed seed otherwise; split indices are hashed and released. Normalization statistics, vocabularies and categorical encodings were fitted on training data only. Short and full training, early stopping and every selection used the validation split. The test split was locked by a software firewall that raises an error unless it is opened after the protocol tag, and it was used once, for the final evaluation of all models. Before the confirmation, the configuration of QuaNAS, the library rule, the seeds (3000--3029), the endpoints, the statistical tests and the claim criteria were committed and tagged in version control, and the search code refuses confirmation tasks without that tag. Because the tag lives in our own repository rather than in a public registry, we call this a \emph{frozen protocol}; the frozen-protocol documents and configurations are released with the code. Pool construction, quality control and baseline training (validation only) do not depend on any search method and were completed before the first confirmation search. At the tag, QuaNAS met the specified search criteria on all 35 development pools. Two decisions were taken after the tag, both before any confirmation search and without access to any accuracy: the fallback chains were extended with fallbacks listed for other slots when a first fallback also failed, and a sensitivity analysis of the time endpoint that excludes two pools timed under external CPU load was added.

QuaNAS was developed in four such cycles; in each, a version was designed on development data, frozen and run once on new datasets, which then became development data. The first version (D1) was a quantum-inspired evolutionary search with reliability-gated proxy fusion, the second (D2) added adaptive proxy fusion, and the third (D3) introduced the transferred priors and the GP search. The present version changed D3 in three steps on the 35 development pools: the library defined by the search space instead of the detected modality, a kernel without main-effect terms, and the reliability switch (mean development AUC 0.0906, 0.0861, 0.0803 and 0.0751 with five seeds).

\subsubsection*{Compared methods}\label{sec:protocol:methods}

All methods search the same pools with the same budget of $B=100$ distinct queries, the same limit of 2,000 proxy evaluations, the same 30 seeds and common random numbers for initial designs. The frozen protocol specified 13 comparators. The five primary ones are \emph{RoBoT-style}, our re-implementation of RoBoT~\cite{he2024robot}, which spends half of the budget on a GP (Mat\'ern-5/2, upper confidence bound) over the weights of a linear combination of min--max normalized proxies, each weight vector querying the best of up to 1,000 scored candidates, and then queries greedily with the best weights; proxy top-$k$ by mean rank (\emph{PTK-MR}), which ranks a uniform sample of up to 2,000 candidates by the mean rank of all proxies and queries the top~\cite{abdelfattah2021zc,krishnakumar2022suitezero}; \emph{QIEA-ZC}, a quantum-inspired evolutionary algorithm~\cite{han2002qiea} with zero-cost proxies that sends four of every five queries to the candidates ranked highest by an equal-weight proxy fusion; regularized evolution (\emph{REA}, population 20, tournaments of 5)~\cite{real2019rea}; and local search (\emph{LS}, restarting from the best of ten random queries)~\cite{white2021ls}. The other eight are random search (RS), QIEA without proxies, a genetic algorithm and population-based incremental learning with reliability-aware proxy fusion (GA-RAPF, PBIL-RAPF), and ranking by a single proxy (SynFlow, NASWOT, SNIP or parameter count). The earlier versions D1--D3 were run alongside, giving 17 search methods.

\textbf{Post-hoc transfer comparators.} The 13 comparators do not use earlier searches. After the confirmation we therefore added four comparators that receive exactly the library of QuaNAS. They use the GP of QuaNAS (kernel $\sigma_c^2+\sigma_{\mathrm{H}}^2e^{-\theta H}$, same hyperparameters, rank-normal targets, expected improvement), so that they differ from it only in how the library is used: \emph{GP-BO}, without transfer, starts from ten random queries; \emph{WS-GP} warm-starts GP-BO~\cite{feurer2015metalearning} with the best architectures of the library tasks, nearest task first by label-free meta-features, each mapped to the nearest unqueried pool member in Hamming distance; \emph{RGPE}~\cite{feurer2018rgpe} combines one GP per library task (at most 2,000 observations each) with a GP on the new task, weighted by the posterior probability of having the lowest ranking loss on the new task's evaluations (256 samples, leave-one-out for the target model, with the weight-dilution rule of the original method), and starts at the optimum of the equally weighted library models; and \emph{NT} queries the pool in the order predicted by the GP of the nearest library task. Their settings were fixed before they were run on the confirmation pools and checked only on the development pools. They are reported separately as post-hoc comparisons.

\textbf{Model-level comparators.} For the comparison of final models, QuaNAS contributes its selected model (see ``\nameref{sec:method:final}'') and four comparators (RoBoT-style, PTK-MR, QIEA-ZC and REA) contribute their best low-fidelity architecture after 50 queries, taking the most frequent one over the 30 seeds.

\subsubsection*{Training, reference models and hardware}\label{sec:protocol:training}

Training recipes were fixed per modality before the pools were built and are identical for all architectures and baselines of a task: AdamW with learning rate $4\times10^{-3}$ and bfloat16 mixed precision; for images and time series a weight decay of 0.05, a cosine schedule with one warm-up epoch, label smoothing 0.1, gradient clipping at 5 and batches of 256 (images) or 128 (time series); for text a weight decay of 0.01, a constant learning rate and batches of 512; for tabular data a weight decay of $10^{-4}$, a constant learning rate, batches of 256 and early stopping with patience 20 in full training. Each confirmation task has a \emph{reference model} trained with 10 seeds under the same recipe: ResNet-18 with a CIFAR-style stem for images~\cite{he2016resnet}, ResNet1D-18 for time series~\cite{wang2017tsc}, CatBoost with the best of 20 random configurations for tabular data~\cite{prokhorenkova2018catboost}, as is standard for boosted trees, and for text the strongest of fastText, fastText with hashed bigrams and TextCNN~\cite{joulin2017fasttext,kim2014textcnn} by mean test accuracy, which makes the text comparison conservative. The references are matched-recipe baselines rather than state-of-the-art models; for instance, Imagenette is used at $32\times32$ pixels. MobileNetV3-Small, EfficientNet-B0, XGBoost, LightGBM and a two-layer multilayer perceptron (MLP) were trained with 3 seeds as further context. The full-training plan comprised the final QuaNAS model (10 seeds), its counterpart from the light-model search, the comparators' architectures (3 seeds each), the zero-query transfer architectures (5 seeds) and the five best pool members (fidelity check): 454 runs and 19.3 GPU-hours, below the specified cap of 40.

All training ran on one NVIDIA GeForce RTX 4080 (16\,GB). The searches themselves are table lookups and ran on the CPU of the same machine (AMD Ryzen 9 9950X, 16 cores). For every trained model we report test accuracy, serialized size in bytes (comparable across neural networks and boosted trees) and CPU latency at batch size 1 on a single thread of this CPU (median of 200 runs after 50 warm-up runs).

\subsubsection*{Metrics and statistical analysis}\label{sec:protocol:stats}

QuaNAS is deterministic up to tie-breaking: its seed-to-seed standard deviation of AUC was zero on 9 of the 12 tasks and at most 0.051 on the others. Seeds are therefore not independent replicates, and the \emph{task} is the unit of replication. For each method and task we average the 30 seeds and compare methods over the 12 task means with one-sided Wilcoxon signed-rank tests (zero differences handled by Pratt's method), Holm-adjusted over the five primary comparators; with 12 tasks the smallest attainable adjusted $p$ is 0.0012. Effect sizes are mean paired differences over tasks with 95\% bootstrap confidence intervals (10,000 resamples). Wins, ties and losses use a region of practical equivalence (ROPE) of $\pm0.01$ AUC~\cite{benavoli2017bayesian}, and QuaNAS and the 13 comparators are ranked with the Friedman test and the Nemenyi critical difference~\cite{demsar2006}. Final models are compared with the reference model by one-sided Welch tests for superiority and for non-inferiority with a margin of 0.5\,pp~\cite{lakens2017tost}, at $\alpha=0.05$. The frozen protocol specified seven hypotheses with explicit criteria, whose outcomes are summarized in the ``\nameref{sec:res:hyp}'' section; sensitivity and acquisition analyses were specified as supplementary analyses without claims. Post-hoc comparisons use the same tests, Holm-adjusted within their own family. For the tail of the task distribution they also report the conditional value at risk $\mathrm{CVaR}_{0.2}$, the mean AUC over the worst 20\% of tasks, compared between two methods by a paired bootstrap interval and a paired permutation test (10,000 resamples each), and they count \emph{negative transfer}, tasks on which a method that uses the library has an AUC more than 0.01 above that of the same GP without the library (GP-BO). Statistical analyses were carried out in Python with SciPy and NumPy; the versions (Python 3.11, SciPy 1.17.1, NumPy 2.4.6) are pinned in the released code. All tests used a significance level of $\alpha=0.05$.

\section*{Results}\label{sec:results}

\subsection*{Evaluation integrity and dataset characteristics}\label{sec:res:integrity}

All 12 confirmation pools passed quality control, with seed-to-seed rank stability between 0.57 (Credit Default) and 0.98 (Chess KRK) (\cref{tab:datasets}); the Amazon Polarity pool contained one diverged training run, which was kept and reported. The modality detector, an operational component, assigned all 39 datasets of the project and all six synthetic edge cases correctly. All 16,920 confirmation searches completed and are reported, every baseline checkpoint reproduced its recorded validation accuracy after reloading, and re-running all 360 QuaNAS searches from the released pools and code on the original hardware reproduced every recorded AUC to within $10^{-14}$.

The confirmation tasks are not easy for proxies (\cref{tab:pools}). The best single proxy reached a Kendall correlation with short-training accuracy between 0.06 and 0.55, it was a different proxy on different tasks, on five tasks (all text and two tabular tasks) even the best proxy reached at most 0.30, and on five tasks some proxy was negatively correlated (down to $-0.39$ on Amazon Polarity). The meta-learned weights of the transferred architecture prior were positive in all three search spaces ($\mu_a=0.56$ for the convolutional, 0.75 for the text and 0.09 for the tabular space), while the proxy prior received small weights ($\mu_s=-0.07$, 0.01 and 0.06). The reliability switch therefore did not trigger on any confirmation task; during development it had triggered on three pools.

\begin{table}[!ht]
\centering
\caption{\textbf{Characteristics of the confirmation pools.}}
\label{tab:pools}
\scriptsize
\setlength{\tabcolsep}{2.6pt}
\begin{tabular}{@{}lcclccc@{}}
\toprule
 & & & & & \multicolumn{2}{c}{Fidelity $r_s$} \\
\cmidrule(lr){6-7}
Dataset & $y^\star$ (\%) & $y^\star-\tilde y$ & Best proxy $\tau$ & Worst $\tau$ & Val. & Test \\
\midrule
OCTMNIST & 90.36 & 1.43 & $+0.54$ (NASWOT) & $+0.08$ & $+0.10$ & $+0.15$\\
TissueMNIST & 61.99 & 2.25 & $+0.52$ (SNIP) & $+0.27$ & $+0.30$ & $+0.50$\\
EMNIST-Balanced & 88.98 & 1.06 & $+0.55$ (SNIP) & $+0.24$ & $-0.67$ & $-0.67$\\
Imagenette-32 & 77.91 & 6.03 & $+0.44$ (NASWOT) & $+0.08$ & $+0.20$ & $+0.20$\\
ElectricDevices & 91.34 & 2.17 & $+0.31$ (NASWOT) & $+0.26$ & $+0.43$ & $-0.63$\\
FetalECG & 94.05 & 7.81 & $+0.46$ (SNIP) & $+0.20$ & $+0.47$ & $-0.43$\\
Yahoo Answers & 67.06 & 8.32 & $+0.26$ (Params) & $-0.39$ & $+0.10$ & $+0.31$\\
Amazon Polarity & 92.02 & 3.06 & $+0.23$ (Params) & $-0.39$ & $-0.10$ & $-0.60$\\
20 Newsgroups & 62.62 & 15.06 & $+0.30$ (Params) & $-0.23$ & $+0.70$ & $+0.20$\\
Chess KRK & 87.35 & 14.40 & $+0.49$ (FLOPs) & $+0.16$ & $+0.87$ & $+0.80$\\
Grid Stability & 97.07 & 3.00 & $+0.06$ (SynFlow) & $-0.21$ & $+0.67$ & $+0.67$\\
Credit Default & 82.58 & 0.62 & $+0.06$ (GradNorm) & $-0.13$ & -- & --\\
\bottomrule
\end{tabular}
\begin{flushleft}\footnotesize $y^\star$, $\tilde y$: best and median short-training accuracy of the pool (difference in pp). $\tau$: Kendall correlation of a proxy with short-training accuracy. Fidelity $r_s$: Spearman correlation between short-training accuracy and full-training validation or test accuracy among the five best pool members (one seed; ``--'': undefined because of ties).
\end{flushleft}
\end{table}

As in any low-fidelity search, short and full training do not rank the very best candidates identically (\cref{tab:pools}, last two columns): among the five best pool members, the rank correlation between short-training and full-training accuracy ranged from strongly positive to negative. With five architectures whose accuracies differ by fractions of a point, none of these correlations is statistically distinguishable from zero, but they show that short training cannot order the top of a pool reliably. This is why all search endpoints below are defined on the low-fidelity objective and the final model is chosen by full-training validation accuracy among a shortlist (see ``\nameref{sec:method:final}'').

\subsection*{Search efficiency on the held-out datasets}\label{sec:res:search}

\begin{figure}[!ht]
\centering
\includegraphics[width=\linewidth]{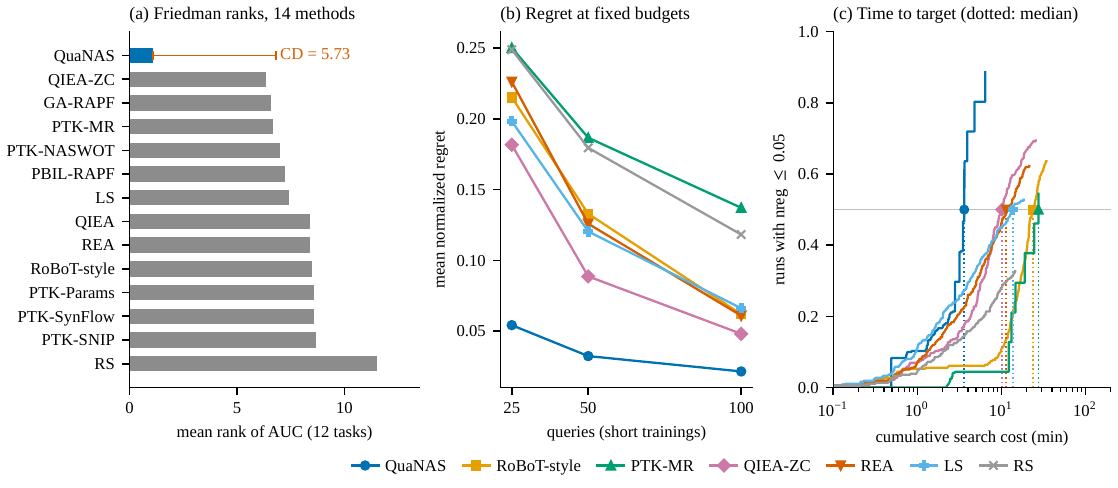}
\caption{\textbf{Search results on the 12 confirmation datasets.} (a) Mean Friedman rank of AUC among QuaNAS and the 13 comparators; the red bar marks the Nemenyi critical difference (CD). (b) Mean normalized regret after 25, 50 and 100 queries. (c) Fraction of the 360 runs per method (12 tasks $\times$ 30 seeds) that have reached a normalized regret of at most 0.05, against the cumulative search cost; markers and dotted lines show the median time of methods that reach the target in at least half of their runs. Method abbreviations as in the section ``Compared methods''; PTK-X: ranking by the single proxy X; QIEA: QIEA without proxies.}
\label{fig:main}
\end{figure}

\Cref{tab:main} and \cref{fig:main} answer the main question. QuaNAS attained a mean anytime regret of 0.058, against 0.155 for QIEA-ZC, 0.177 for local search, 0.182 for RoBoT-style, 0.184 for REA and 0.226 for proxy ranking, i.e.\ 2.7 to 3.9 times lower. It had the lower regret on every one of the 12 tasks against each of the five primary comparators; with the equivalence region of $\pm0.01$, this amounts to 11 or 12 wins, at most one tie and no loss per comparator. The mean paired differences ranged from 0.097 (QIEA-ZC) to 0.167 (PTK-MR), all 95\% confidence intervals exclude zero, and all Holm-adjusted $p$ equal 0.0012, the smallest value attainable with 12 tasks. Among QuaNAS and the 13 comparators, QuaNAS ranked first on 11 of 12 tasks (mean rank 1.08; next QIEA-ZC 6.33; Friedman $p=1.5\times10^{-5}$, critical difference 5.73).

\begin{table}[!ht]
\centering
\caption{\textbf{Search efficiency on the 12 confirmation datasets against the pre-specified comparators.}}
\label{tab:main}
\footnotesize
\setlength{\tabcolsep}{4.2pt}
\begin{tabular}{@{}lccccccc@{}}
\toprule
 & QuaNAS & RoBoT-style & PTK-MR & QIEA-ZC & REA & LS & RS \\
\midrule
\multicolumn{8}{@{}l}{\emph{Anytime regret} (AUC, lower is better)}\\
OCTMNIST & \textbf{0.000} & 0.130 & 0.056 & 0.098 & 0.185 & 0.180 & 0.247\\
TissueMNIST & \textbf{0.010} & 0.208 & 0.114 & 0.149 & 0.249 & 0.241 & 0.298\\
EMNIST-Balanced & \textbf{0.021} & 0.068 & 0.023 & 0.079 & 0.107 & 0.108 & 0.143\\
Imagenette-32 & \textbf{0.000} & 0.321 & 0.152 & 0.187 & 0.281 & 0.237 & 0.320\\
ElectricDevices & \textbf{0.032} & 0.103 & 0.058 & 0.163 & 0.225 & 0.228 & 0.280\\
FetalECG & \textbf{0.079} & 0.159 & 0.091 & 0.133 & 0.146 & 0.152 & 0.184\\
Yahoo Answers & \textbf{0.037} & 0.195 & 0.543 & 0.144 & 0.125 & 0.141 & 0.171\\
Amazon Polarity & \textbf{0.055} & 0.221 & 0.695 & 0.145 & 0.124 & 0.127 & 0.149\\
20 Newsgroups & \textbf{0.012} & 0.135 & 0.214 & 0.133 & 0.120 & 0.117 & 0.145\\
Chess KRK & \textbf{0.090} & 0.190 & 0.133 & 0.137 & 0.164 & 0.123 & 0.218\\
Grid Stability & \textbf{0.146} & 0.233 & 0.282 & 0.214 & 0.190 & 0.198 & 0.246\\
Credit Default & \textbf{0.216} & 0.226 & 0.345 & 0.278 & 0.292 & 0.275 & 0.291\\
\midrule
Mean AUC & \textbf{0.058} & 0.182 & 0.226 & 0.155 & 0.184 & 0.177 & 0.224\\
Mean rank (14 methods) & \textbf{1.08} & 8.50 & 6.67 & 6.33 & 8.42 & 7.42 & 11.50\\
$\Delta\mathrm{AUC}$ vs.\ QuaNAS & -- & $+0.124$ & $+0.167$ & $+0.097$ & $+0.126$ & $+0.119$ & $+0.166$\\
95\% CI & -- & [0.084, 0.172] & [0.073, 0.289] & [0.075, 0.121] & [0.087, 0.170] & [0.082, 0.159] & --\\
W/T/L (ROPE 0.01) & -- & 11/1/0 & 11/1/0 & 12/0/0 & 12/0/0 & 12/0/0 & --\\
Holm-adjusted $p$ & -- & 0.0012 & 0.0012 & 0.0012 & 0.0012 & 0.0012 & --\\
Development, 35 pools & \textbf{0.075} & 0.212 & 0.215 & 0.188 & 0.203 & 0.189 & 0.245\\
\midrule
\multicolumn{8}{@{}l}{\emph{Time to target} ($\mathrm{TTT}_{0.05}$, low-fidelity objective)}\\
Mean $\mathrm{TTT}_{0.05}$ (s) & \textbf{197} & 1042 & 1068 & 589 & 677 & 817 & 1001\\
Runs reaching target (\%) & \textbf{88.6} & 63.6 & 54.4 & 69.4 & 62.2 & 52.8 & 32.8\\
Wins/losses (TTT) & -- & 11/1 & 12/0 & 10/2 & 10/2 & 9/3 & --\\
Holm-adjusted $p$ (TTT) & -- & 0.0029 & 0.0012 & 0.0037 & 0.0068 & 0.0068 & --\\
\bottomrule
\end{tabular}
\begin{flushleft}\footnotesize AUC: \cref{eq:auc}, mean of 30 seeds; best per row in bold. $\Delta\mathrm{AUC}$: mean over tasks of comparator minus QuaNAS with a 95\% bootstrap interval. W/T/L: tasks on which QuaNAS is better, equivalent ($|\Delta|\le0.01$) or worse. $p$: one-sided Wilcoxon signed-rank tests over the 12 task means, Holm-adjusted over the five primary comparators (smallest attainable value 0.0012). Mean rank: Friedman rank among QuaNAS and the 13 comparators. $\mathrm{TTT}_{0.05}$: \cref{eq:ttt}; runs that do not reach the target are censored. Development: mean AUC on the 35 development pools.
\end{flushleft}
\end{table}

The advantage appears from the first queries (\cref{fig:main}b). After 25 queries the mean normalized regret of QuaNAS was 0.054, against 0.182--0.250 for the primary comparators, and it had already found the best pool member on 7 tasks. On OCTMNIST and Imagenette-32 the very first query, the prior optimum, was the best of the 600 candidates, an architecture that lies in expectation in the top 0.17\% of the convolutional space (see ``\nameref{sec:problem}''). After 100 queries QuaNAS still had the lowest regret (0.022 versus 0.048--0.137). The development pools, on which the configuration had been frozen, show the same picture (\cref{tab:main}; Holm $p\le1.1\times10^{-8}$ against each primary comparator). Defined as confirmation minus development regret, the generalization gap of QuaNAS was $-0.016$, within the range of the primary comparators ($-0.033$ to $+0.011$): moving from the data used for design to new datasets and new architectures did not erode its advantage.

\subsection*{Search cost and time to a near-optimal architecture}\label{sec:res:cost}

For a practitioner the relevant cost is the compute spent until a near-optimal architecture has been found. QuaNAS needed on average 197\,s of short training, proxy computation and overhead to reach a normalized regret of 0.05, against 589\,s (QIEA-ZC), 677\,s (REA), 817\,s (LS), 1042\,s (RoBoT-style) and 1068\,s (PTK-MR), i.e.\ 3.0 to 5.4 times faster on this low-fidelity objective (\cref{tab:main}). The geometric means of the per-task speed-ups ranged from 2.6 (QIEA-ZC; 95\% CI 1.6--4.5) to 4.6 (PTK-MR; 2.7--8.0), it was faster on 9 to 12 of the 12 tasks against each comparator, and all Holm-adjusted $p\le0.0068$. Within the budget, 88.6\% of QuaNAS runs reached the target, against 52.8--69.4\% for the primary comparators (\cref{fig:main}c). Two pools (Amazon Polarity, FetalECG) were built while other users loaded the shared CPU, which lengthened part of their recorded training times; excluding them leaves all five comparisons significant (Holm $p=0.0049$ against PTK-MR and $p=0.0078$ against the other four).

QuaNAS was not the fastest on every task. On the two cheapest tabular tasks (Chess KRK and Grid Stability, where all six methods reached the target within 55--172\,s on average) and on FetalECG, local search, REA or RoBoT-style got there first. Over the \emph{full} budget of 100 queries, QuaNAS spent 914\,s per search, more than random search (602\,s), local search (660\,s) and REA (695\,s) and less than QIEA-ZC (981\,s), RoBoT-style (1255\,s) and PTK-MR (1356\,s). This is the price of concentrating on strong candidates, which tend to be larger and slower to train, and of computing proxies for the screened fifth of the pool and for every queried architecture (294 proxy evaluations per search on average, against 700 for RoBoT-style and 950 for PTK-MR). The overhead of the GP itself, 0.28\,s per search, is negligible. QuaNAS therefore pays off when a good architecture is needed early, the purpose of query-efficient search; when the whole budget is spent anyway, local search costs less per query but ends at a higher regret (see ``\nameref{sec:res:search}'').

\subsection*{Comparison with transfer-based Bayesian optimization}\label{sec:res:transfer-bo}

The pre-specified comparators do not use earlier searches, so part of the advantage of QuaNAS could come from the library alone. \Cref{tab:posthoc} therefore compares QuaNAS with the four post-hoc comparators that receive the same library and share its GP (see ``\nameref{sec:protocol:methods}''). Without transfer, the same GP reached a mean AUC of 0.130 (GP-BO), better than every pre-specified comparator but more than twice that of QuaNAS (Holm $p=0.0020$; QuaNAS better on 11 tasks and tied on one). Ranking the pool by the GP of the single nearest library task, without any feedback (NT), already reached 0.108. The library, not the surrogate, is therefore the main source of efficiency.

\begin{table}[!ht]
\centering
\caption{\textbf{Post-hoc comparison with methods that use the same library and the same Gaussian process.}}
\label{tab:posthoc}
\scriptsize
\setlength{\tabcolsep}{2.6pt}
\begin{adjustwidth}{-0.55in}{0in}
\begin{tabular}{@{}lccccccccccc@{}}
\toprule
 & \multicolumn{8}{c}{Confirmation (12 held-out datasets)} & \multicolumn{3}{c}{Development (35 pools)} \\
\cmidrule(lr){2-9}\cmidrule(lr){10-12}
Method & AUC & Median & $\mathrm{CVaR}_{0.2}$ & Neg. & $\Delta\mathrm{AUC}$ [95\% CI] & W/T/L & Holm $p$ & $\mathrm{TTT}_{0.05}$ (s) & AUC & $\mathrm{CVaR}_{0.2}$ & Neg. \\
\midrule
QuaNAS & \textbf{0.058} & 0.034 & \textbf{0.151} & \textbf{0} & -- & -- & -- & 197 & \textbf{0.075} & \textbf{0.208} & \textbf{2}\\
\midrule
RGPE~\cite{feurer2018rgpe} & \textbf{0.058} & \textbf{0.020} & 0.174 & 1 & $+0.000$ [$-0.012$, 0.013] & 3/6/3 & 0.62 & \textbf{146} & 0.086 & 0.274 & 5\\
WS-GP~\cite{feurer2015metalearning} & 0.075 & 0.050 & 0.154 & 1 & $+0.017$ [$-0.003$, 0.040] & 5/4/3 & 0.38 & 180 & 0.097 & 0.296 & 5\\
NT (nearest task) & 0.108 & 0.058 & 0.291 & 2 & $+0.050$ [0.005, 0.112] & 6/6/0 & 0.028 & 262 & 0.160 & 0.457 & 11\\
GP-BO (no transfer) & 0.130 & 0.120 & 0.208 & -- & $+0.072$ [0.044, 0.101] & 11/1/0 & 0.0020 & 299 & 0.154 & 0.300 & --\\
\bottomrule
\end{tabular}
\end{adjustwidth}
\begin{flushleft}\footnotesize All methods receive the library of QuaNAS and use its GP (kernel, hyperparameters, rank-normal targets, expected improvement); their settings were fixed before they were run on the confirmation pools. AUC: mean over tasks; median and $\mathrm{CVaR}_{0.2}$ (mean of the worst 20\% of tasks) over tasks; lowest value per column in bold. Neg.: number of tasks with negative transfer (AUC more than 0.01 above GP-BO). $\Delta\mathrm{AUC}$: comparator minus QuaNAS; W/T/L as in \cref{tab:main}; $p$: one-sided Wilcoxon tests for ``QuaNAS lower'', Holm-adjusted over the four comparators. Development: 10 seeds per pool for the comparators; QuaNAS was configured on these pools.
\end{flushleft}
\end{table}

Among the methods that combine the library with feedback from the new task, QuaNAS matched the strongest, RGPE, in mean regret (0.058 and 0.058; mean difference 0.000, 95\% CI $[-0.012, 0.013]$; 3 wins, 6 ties and 3 losses for QuaNAS; Holm-adjusted one-sided $p=0.62$ for a lower regret of QuaNAS, two-sided $p=0.82$), and warm-started GP-BO came next (0.075; Holm $p=0.38$). The two methods differ in the shape of their results. RGPE, whose per-task GPs carry interactions between decisions that the additive main-effect prior of QuaNAS cannot express, had the lower median regret (0.020 versus 0.034). Across all 47 pools the two methods won in different spaces: RGPE on the 34 image, time-series and text pools (mean AUC 0.034 versus 0.043; better on 14 and worse on 5) and QuaNAS on the 13 tabular pools (0.142 versus 0.199; better on 9 and worse on 4). The largest losses of RGPE relative to QuaNAS occurred on tabular tasks (Credit Default and Grid Stability; Adult, Dry Bean and MAGIC among the development pools), the space in which earlier searches transfer least ($\mu_a=0.09$; the ``\nameref{sec:res:transfer}'' section) and in which the re-estimated prior weights, the proxy prior and, during development, the reliability switch protect QuaNAS.

These losses give QuaNAS the lighter tail. Its $\mathrm{CVaR}_{0.2}$ was 0.151 against 0.174 for RGPE, a difference that 12 tasks cannot resolve (permutation $p=0.18$). No negative transfer was observed for QuaNAS on the held-out datasets, against one dataset each for RGPE (Grid Stability) and WS-GP (Chess KRK) and two for NT, too few events for a statistical comparison. Over all 47 pools the tail difference was larger ($\mathrm{CVaR}_{0.2}$ 0.196 versus 0.253, 95\% CI of the difference $[0.003, 0.119]$, $p=0.032$; negative transfer on 2 versus 6 pools), but the development pools favor QuaNAS, whose configuration was selected on them. RGPE, in turn, reached the near-optimal target sooner on average (146 versus 197\,s; sooner on 10 of 12 tasks, two-sided $p=0.11$): it computes no proxies, whereas QuaNAS spends 83\,s per search on average, and up to 191\,s on image tasks, scoring the screened fifth of the pool before its first query. Without its proxy prior, QuaNAS reached the target in 133\,s, at a higher regret (0.079).

\subsection*{Which components matter?}\label{sec:res:ablation}

\begin{figure}[!ht]
\centering
\includegraphics[width=\linewidth]{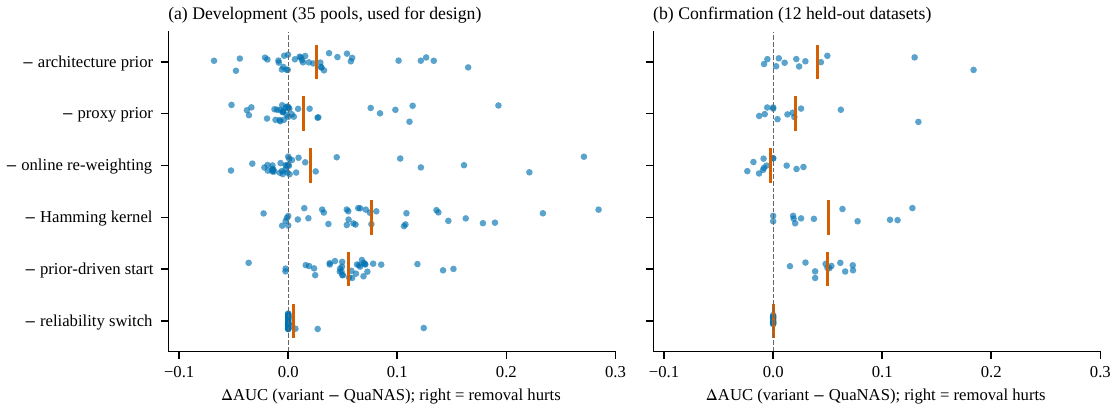}
\caption{\textbf{Change of AUC when one component is removed (each dot is a pool or dataset; red bars are means).} Points to the right of zero are tasks on which the removal hurts.}
\label{fig:ablation}
\end{figure}

\Cref{tab:ablation} and \cref{fig:ablation} remove one component at a time, on the development pools and on the confirmation datasets. Three components were necessary in both sets. Replacing the prior-driven start by ten random queries raised regret on all 12 confirmation tasks ($\Delta=+0.050$, Holm $p=0.0015$); removing the Hamming kernel raised it by 0.051 ($p=0.0049$); and removing the transferred architecture prior raised it by 0.041 ($p=0.019$), with losses of up to 0.18 (Imagenette-32). Knowledge from earlier searches, used both as a prior and as the starting point, therefore accounts for a large part of the gains, in agreement with the ``\nameref{sec:res:transfer-bo}'' section, and it pays off from the first query.

\begin{table}[!ht]
\centering
\caption{\textbf{Component removals on the development pools and the confirmation datasets.}}
\label{tab:ablation}
\footnotesize
\setlength{\tabcolsep}{4.5pt}
\begin{tabular}{@{}lcccccccc@{}}
\toprule
 & \multicolumn{4}{c}{Development (35 pools)} & \multicolumn{4}{c}{Confirmation (12 held-out datasets)} \\
\cmidrule(lr){2-5}\cmidrule(lr){6-9}
Removed component & AUC & $\Delta$ & worst & $p$ & AUC & $\Delta$ & worst & $p$ \\
\midrule
QuaNAS (full configuration) & 0.075 & -- & -- & -- & 0.058 & -- & -- & --\\
\midrule
\textendash\ architecture prior $a$ (mean and kernel) & 0.100 & $+0.026$ & $+0.165$ & 0.013 & 0.099 & $+0.041$ & $+0.184$ & 0.019\\
\textendash\ proxy prior $s$ (no proxy computed) & 0.089 & $+0.014$ & $+0.193$ & 0.99 & 0.079 & $+0.021$ & $+0.133$ & 0.13\\
\textendash\ online re-weighting ($\sigma_a=\sigma_s=0$) & 0.095 & $+0.021$ & $+0.271$ & 0.99 & 0.056 & $-0.002$ & $+0.028$ & 1.00\\
\textendash\ Hamming kernel ($\sigma_{\mathrm{H}}=0$) & 0.151 & $+0.076$ & $+0.285$ & $4.1\times10^{-6}$ & 0.109 & $+0.051$ & $+0.128$ & 0.0049\\
\textendash\ prior-driven start (10 random queries) & 0.130 & $+0.055$ & $+0.151$ & $7.5\times10^{-9}$ & 0.108 & $+0.050$ & $+0.073$ & 0.0015\\
\textendash\ reliability switch & 0.079 & $+0.005$ & $+0.124$ & 0.13 & \multicolumn{4}{c}{identical: switch not triggered}\\
\bottomrule
\end{tabular}
\begin{flushleft}\footnotesize $\Delta$: mean change (variant minus QuaNAS; positive means the removal hurts). Worst: largest single-task increase. $p$: Holm-adjusted one-sided Wilcoxon test for ``the removal increases AUC'' over tasks; the opposite tests gave Holm $p=1$ for every component in both sets.
\end{flushleft}
\end{table}

The other three components act on rare, large failures rather than on the mean, so their benefit does not show in the mean over the 12 confirmation datasets. Without the proxy prior, regret rose by 0.021 on average ($p=0.13$) and by 0.13 on Credit Default, with no task improving by more than 0.013. Fixing the prior weights at their meta-learned values ($\sigma_a=\sigma_s=0$) left mean regret on the confirmation datasets unchanged ($-0.002$; three tasks better, three worse), because the meta-learned weights were adequate there. On the development pools, however, where some weights were wrong, fixing them increased regret by up to 0.27 on a single pool (Adult) while gaining at most 0.05 elsewhere. The reliability switch was inactive on the confirmation datasets and, on the three development pools where it fired, prevented losses of up to 0.12. Online re-weighting and the switch are therefore safeguards against rare failures, whose benefit was observed on the development pools. The development versions show the same pattern: the transferred priors and the GP search introduced with the third version (D3) brought the main gain, and D3, which lacked the last three changes (see ``\nameref{sec:protocol:prereg}''), already reached a similar mean AUC on the confirmation datasets (0.073; 3 wins, 7 ties and 2 losses for the final version; two-sided $p=0.36$), whereas the first two versions reached 0.149 (D2) and 0.173 (D1). No removal improved QuaNAS significantly in either set (all Holm $p=1$); because fixed weights had a marginally lower confirmation mean (0.056 versus 0.058, not significant), the frozen hypothesis on components is partly supported.

Two further analyses were specified without claims. Replacing expected improvement by the posterior mean, a greedy acquisition, raised the confirmation regret to 0.077 (two-sided $p=0.018$), with five tasks worse and none better; during development, greedy acquisition had won on more pools but lost up to 0.14 on others, which is why expected improvement was kept. Setting the screening fraction to 0.1 or 0.3, or halving or doubling the Hamming rate or the noise variance, increased mean regret to 0.065--0.086 on the confirmation datasets and to 0.080--0.091 on the development pools, so the defaults, fixed during development, were the best of those tested in both sets and need no per-task tuning. Single tasks can still react strongly to a change: with half the screening fraction ($\rho=0.1$), regret rose by up to 0.174 on one confirmation dataset.

\subsection*{Robustness: library composition, manipulated libraries and a full search space}\label{sec:res:robust}

Several confirmation datasets come from collections that also supplied development pools, which might make the transfer easier than it would be for an unrelated dataset. We therefore re-ran QuaNAS with reduced libraries, unchanged otherwise. In a leave-family-out library we removed, for each confirmation dataset, every library pool from the same collection or problem family (MedMNIST for OCTMNIST and TissueMNIST; Fashion-MNIST and KMNIST for EMNIST; CIFAR-10 and Spherical CIFAR-100 for Imagenette; FordA and Satellite for the two time-series tasks; AG News, DBPedia and Yelp Polarity, from the collection of Zhang et al.~\cite{zhang2015charcnn}, for Yahoo Answers and Amazon Polarity, plus IMDB for the latter; AG News and DBPedia for 20 Newsgroups; and Connect-4, MAGIC and Adult for Chess KRK, Grid Stability and Credit Default). Mean AUC rose from 0.058 to 0.081. The image, time-series and text tasks were almost unaffected (largest change 0.016), whereas the tabular tasks, whose libraries shrank from six to five pools, lost most (Credit Default 0.357 instead of 0.216, Chess KRK 0.168 instead of 0.090). In a stricter variant we removed the third of each library whose own main-effect model agreed best with the target's accuracies, a choice that uses the target labels and is therefore possible only in analysis; mean AUC rose to 0.078. In both variants QuaNAS still had lower regret than each primary comparator (leave-family-out library: Holm $p=0.012$ against REA and $p=0.019$ against the other four; stricter variant: $p=0.0024$ against PTK-MR and $p=0.0012$ against the other four). The transfer thus does not depend on near-duplicate datasets, except that a small tabular library benefits from one related task.

Stress tests with manipulated libraries of the 21 convolutional pools delimit the safeguard of QuaNAS (\nameref{S1_Appendix}, Table A). QuaNAS had a lower mean AUC than RGPE when the accuracies in the library were scrambled (0.120 versus 0.164), which its leave-one-task-out check detects before the first evaluation, and a higher one when the library held only one or two earlier searches (0.103 versus 0.080 and 0.073 versus 0.055) or stemmed only from the other data type (0.108 versus 0.061), a library that is consistent within itself and therefore passes the check.

To test the method on a standard benchmark without pool sampling or a fidelity gap, we also ran every method on the whole NAS-Bench-201 space~\cite{dong2020nb201}, whose 15,625 cells were trained for 200 epochs on CIFAR-10, CIFAR-100 and ImageNet16-120, with the other two datasets as the library and the 200-epoch validation accuracy as the search objective (\cref{tab:nb201}; 10 seeds; the benchmark file provides only two zero-cost proxies, the parameter count and FLOPs). QuaNAS lowered the mean AUC 2.1- to 5.8-fold relative to the methods without transfer (0.049 versus 0.105--0.287) and reached the near-optimal target in every run. The transfer comparators were better still (mean AUC 0.002--0.010), and two of them found the best or a near-best cell within one to three queries, because the three datasets share all 15,625 cells and rank them almost identically (Kendall $\tau=0.81$--0.85): the surrogate of an earlier task is nearly a lookup table for the new one, whereas the additive main effects of QuaNAS do not represent the interactions between edge operations that decide accuracy in this cell space. NAS-Bench-201 thus marks the boundary of main-effect transfer: when earlier tasks share the candidates and their ranking, per-task surrogates are the better transfer model.

\begin{table}[!ht]
\centering
\caption{\textbf{NAS-Bench-201, whole space (post hoc).}}
\label{tab:nb201}
\scriptsize
\setlength{\tabcolsep}{2.4pt}
\begin{tabular}{@{}lccccccc@{}}
\toprule
 & \multicolumn{2}{c}{CIFAR-10} & \multicolumn{2}{c}{CIFAR-100} & \multicolumn{2}{c}{ImageNet16-120} & Mean \\
\cmidrule(lr){2-3}\cmidrule(lr){4-5}\cmidrule(lr){6-7}
Method & AUC & Acc.@25 & AUC & Acc.@25 & AUC & Acc.@25 & AUC \\
\midrule
QuaNAS & 0.060 & 94.00 & 0.025 & 73.18 & 0.063 & 46.07 & 0.049\\
RGPE & \textbf{0.001} & 94.36 & 0.003 & 73.24 & \textbf{0.001} & 47.33 & \textbf{0.002}\\
WS-GP & 0.008 & 94.36 & 0.011 & 73.10 & 0.011 & 47.20 & 0.010\\
NT & \textbf{0.001} & 94.36 & \textbf{0.002} & 73.24 & \textbf{0.001} & 47.33 & \textbf{0.002}\\
\midrule
GP-BO & 0.140 & 93.77 & 0.125 & 72.05 & 0.118 & 45.74 & 0.128\\
REA & 0.240 & 93.34 & 0.212 & 70.51 & 0.179 & 44.73 & 0.210\\
LS & 0.185 & 93.47 & 0.169 & 71.34 & 0.153 & 45.20 & 0.169\\
RS & 0.320 & 93.22 & 0.319 & 70.23 & 0.220 & 44.76 & 0.287\\
RoBoT-style & 0.178 & 93.96 & 0.137 & 72.40 & 0.129 & 45.91 & 0.148\\
PTK-MR & 0.121 & 93.96 & 0.098 & 72.50 & 0.097 & 46.15 & 0.105\\
QIEA-ZC & 0.176 & 93.72 & 0.143 & 71.69 & 0.135 & 46.02 & 0.151\\
\bottomrule
\end{tabular}
\begin{flushleft}\footnotesize Library: the other two datasets (same 15,625 cells). AUC over 100 queries on the 200-epoch validation accuracy (best AUC per column in bold). Acc.@25: accuracy (\%) of the best cell found after 25 queries, CIFAR-10 test accuracy of the cell with the best validation accuracy and, for the other two datasets, the benchmark's evaluation accuracy. Proxies available: parameter count and FLOPs. Upper block: methods that use the library.
\end{flushleft}
\end{table}

\subsection*{Final models: accuracy, size and latency}\label{sec:res:models}

\begin{figure}[!ht]
\centering
\includegraphics[width=0.76\linewidth]{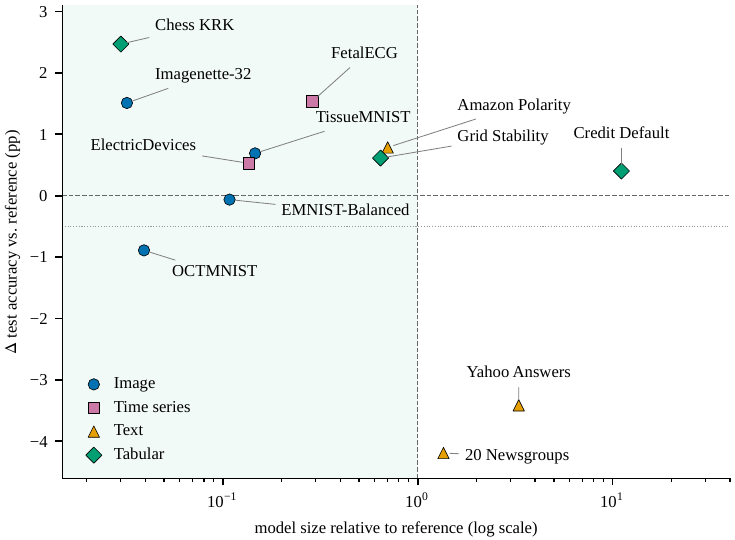}
\caption{\textbf{Accuracy difference of the final QuaNAS models to the reference models against their relative size (log scale).} The shaded region contains models that are smaller than the reference; the dotted line marks the non-inferiority margin of $-0.5$\,pp.}
\label{fig:models}
\end{figure}

\Cref{tab:models} compares the final QuaNAS models (10 seeds) with the reference models (10 seeds), all trained with identical recipes and evaluated once on the locked test splits. QuaNAS was superior on 7 datasets and non-inferior within 0.5\,pp on 2 more, so the frozen criterion (7 of 12) was met with 9 of 12. The largest gains were on Chess KRK (+2.47\,pp over CatBoost), FetalECG (+1.54\,pp over ResNet1D-18) and Imagenette-32 (+1.51\,pp over ResNet-18). On 8 of these 9 datasets the QuaNAS model was also smaller (\cref{fig:models}): image and time-series models had 3--29\% of the size of the ResNet references and ran up to 4.7 times faster on a single CPU thread. On Credit Default the QuaNAS model was more accurate but larger than the very compact CatBoost model. Against the strongest matched baseline of each dataset (one to four per dataset), not only the pre-specified reference, QuaNAS was more accurate on 7 datasets and within 0.5\,pp on 9.

\begin{table}[!ht]
\centering
\caption{\textbf{Final models on the locked test splits.}}
\label{tab:models}
\footnotesize
\setlength{\tabcolsep}{4.5pt}
\begin{tabular}{@{}llccrlcc@{}}
\toprule
Dataset & Reference & Reference acc. & QuaNAS acc. & $\Delta$ (pp) & Outcome & Size ratio & CPU latency (ms) \\
\midrule
OCTMNIST & ResNet-18 & $78.47\pm1.16$ & $77.58\pm0.87$ & $-0.89$ & not met & 0.04 & 7.14 / 1.53\\
TissueMNIST & ResNet-18 & $72.26\pm0.11$ & $72.95\pm0.07$ & $+0.69$ & superior & 0.15 & 7.09 / 2.76\\
EMNIST-Balanced & ResNet-18 & $91.02\pm0.07$ & $90.96\pm0.09$ & $-0.06$ & non-inferior & 0.11 & 7.15 / 1.95\\
Imagenette-32 & ResNet-18 & $83.96\pm0.35$ & $85.47\pm0.34$ & $+1.51$ & superior & 0.03 & 7.96 / 2.61\\
ElectricDevices & ResNet1D-18 & $71.47\pm0.61$ & $72.00\pm0.76$ & $+0.53$ & non-inferior & 0.14 & 2.10 / 0.80\\
FetalECG & ResNet1D-18 & $92.41\pm0.50$ & $93.94\pm0.43$ & $+1.54$ & superior & 0.29 & 4.79 / 4.78\\
Yahoo Answers & fastText & $66.59\pm0.13$ & $63.18\pm0.36$ & $-3.41$ & not met & 3.30 & 0.02 / 18.6\\
Amazon Polarity & fastText-bigram & $89.63\pm0.03$ & $90.42\pm0.31$ & $+0.79$ & superior & 0.70 & 0.04 / 18.5\\
20 Newsgroups & fastText & $64.87\pm0.36$ & $60.68\pm0.29$ & $-4.19$ & not met & 1.36 & 0.02 / 5.58\\
Chess KRK & CatBoost & $88.26\pm0.25$ & $90.73\pm0.80$ & $+2.47$ & superior & 0.03 & 0.06 / 0.09\\
Grid Stability & CatBoost & $94.94\pm0.28$ & $95.55\pm0.47$ & $+0.61$ & superior & 0.64 & 0.05 / 0.05\\
Credit Default & CatBoost & $81.79\pm0.09$ & $82.20\pm0.22$ & $+0.40$ & superior & 11.10 & 0.05 / 0.07\\
\bottomrule
\end{tabular}
\begin{flushleft}\footnotesize Test accuracy in \% (mean$\pm$SD over 10 seeds). Outcome: one-sided Welch tests at $\alpha=0.05$ for superiority and for non-inferiority with a margin of 0.5\,pp. Size ratio: serialized bytes of the QuaNAS model over those of the reference. CPU latency: single thread, batch size 1 (reference / QuaNAS).
\end{flushleft}
\end{table}

The comparison between searches at the model level is not symmetric: QuaNAS contributed the winner of its full-training shortlist, whereas each comparator contributed its most frequent best architecture after 50 queries. The comparators' architectures were close in accuracy (QuaNAS higher by 0.16--0.64\,pp on average): given enough queries, all searches find good members of the same pool, and what distinguishes QuaNAS is how early it finds them (see ``\nameref{sec:res:cost}''). The three datasets on which QuaNAS did not match the reference are analyzed in the ``\nameref{sec:res:errors}'' section.

\subsection*{Knowledge transfer without queries and light models}\label{sec:res:transfer}

To isolate the transferred knowledge, we trained the prior optimum of each confirmation task, $\arg\max_x\mu_a a(x)+\mu_s s(x)$, without any evaluation on that task (\cref{tab:transfer}). It was non-inferior to the reference on 9 of 12 datasets and not worse than the corresponding RoBoT-style transfer (the RoBoT-style architecture of the nearest development dataset under label-free meta-features) on 11; it was the more accurate of the two on 8 datasets (mean difference to the reference $-0.70$ versus $-0.99$\,pp). The earlier searches alone, read through the main-effect models, therefore point to competitive architectures on new datasets.

\begin{table}[!ht]
\centering
\caption{\textbf{Zero-query transfer and light models.}}
\label{tab:transfer}
\footnotesize
\setlength{\tabcolsep}{4.5pt}
\begin{tabular}{@{}lrrcc|rccc@{}}
\toprule
 & \multicolumn{4}{c|}{Zero-query transfer} & \multicolumn{4}{c}{Light model (capped search)} \\
Dataset & QuaNAS & RoBoT-style & Non-inf. & Not worse & $\Delta$ (pp) & Size ratio & Latency ratio & All met \\
\midrule
OCTMNIST & $-0.49$ & $-0.81$ & -- & \checkmark & $-0.89$ & 0.04 & 0.21 & --\\
TissueMNIST & $+0.76$ & $+0.48$ & \checkmark & \checkmark & $+0.49$ & 0.08 & 0.35 & \checkmark\\
EMNIST-Balanced & $+0.03$ & $+0.01$ & \checkmark & \checkmark & $+0.04$ & 0.06 & 0.27 & \checkmark\\
Imagenette-32 & $+1.21$ & $-2.29$ & \checkmark & \checkmark & $+1.51$ & 0.03 & 0.33 & \checkmark\\
ElectricDevices & $+1.20$ & $+1.08$ & \checkmark & \checkmark & $+1.24$ & 0.12 & 0.41 & \checkmark\\
FetalECG & $+1.14$ & $+0.94$ & \checkmark & \checkmark & $+1.28$ & 0.18 & 0.60 & \checkmark\\
Yahoo Answers & $-2.70$ & $-3.23$ & -- & \checkmark & $-3.18$ & 1.36 & 301 & --\\
Amazon Polarity & $+0.57$ & $+0.59$ & \checkmark & \checkmark & $+1.79$ & 1.35 & 351 & --\\
20 Newsgroups & $-12.49$ & $-11.95$ & -- & \checkmark & $-4.19$ & 1.36 & 306 & --\\
Chess KRK & $+1.80$ & $+3.09$ & \checkmark & -- & $+2.47$ & 0.03 & 1.48 & --\\
Grid Stability & $+0.23$ & $+0.27$ & \checkmark & \checkmark & $+0.29$ & 0.44 & 0.60 & --\\
Credit Default & $+0.35$ & $-0.02$ & \checkmark & \checkmark & $+0.37$ & 1.45 & 1.14 & --\\
\bottomrule
\end{tabular}
\begin{flushleft}\footnotesize Left: accuracy difference to the reference (pp) of the transferred prior optimum (QuaNAS, 5 seeds) and of RoBoT-style transfer (5 seeds), and whether QuaNAS transfer is non-inferior to the reference and not worse than RoBoT-style transfer. Right: capped final model, its accuracy difference to the reference, size ratio, CPU latency ratio (model over reference; below 1 is faster) and whether all three light-model conditions hold.
\end{flushleft}
\end{table}

\begin{figure}[!ht]
\centering
\includegraphics[width=0.97\linewidth]{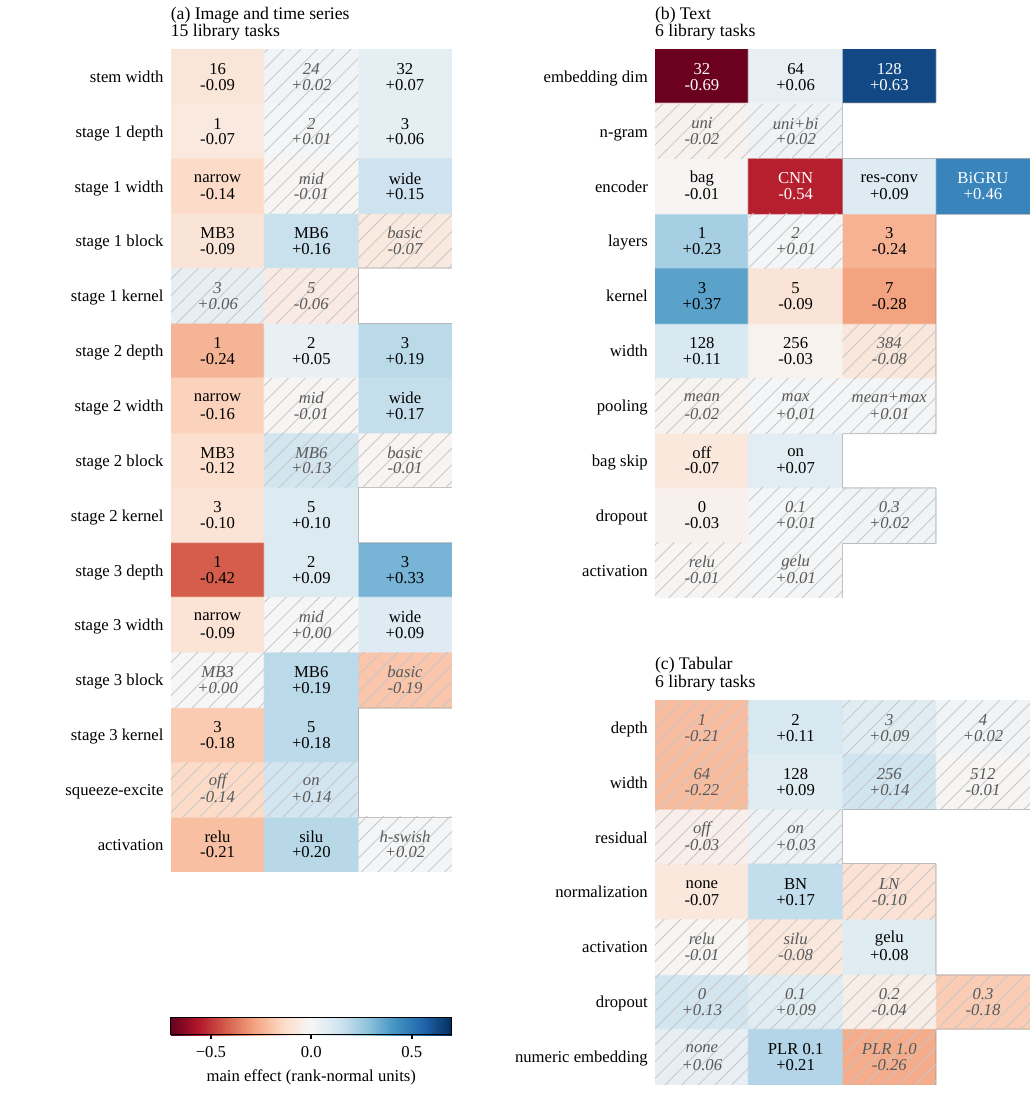}
\caption{\textbf{Knowledge transferred by QuaNAS.} Library mean of the centered ridge main effect of every option (rank-normal units; blue favored, red disfavored) in the three search spaces. Hatched cells in grey italics: fewer than 75\% of the library tasks share the sign of the effect. Every cell prints its value, so the signs remain readable in black and white. Option labels: MB3/MB6, MBConv block with expansion 3/6; basic, residual basic block; h-swish, Hardswish; uni/uni+bi, unigram or unigram plus hashed-bigram input; CNN, TextCNN; res-conv, residual 1-D convolution; BN/LN, batch/layer normalization; PLR 0.1/1.0, periodic numerical embeddings with initialization scale 0.1/1.0.}
\label{fig:effects}
\end{figure}

\Cref{fig:effects} shows what this knowledge consists of. In the convolutional space the library favored deeper second and third stages, kernel size 5 in these stages, MBConv blocks with expansion 6 in the last stage and SiLU activations; in the text space it favored 128-dimensional embeddings, BiGRU encoders and kernel size 3, and disfavored TextCNN encoders. On average 80\% of the library tasks shared the sign of an effect in these two spaces, against 65\% in the tabular space, where the largest effects, such as that of the numeric embedding, changed sign between tasks. This matches the meta-learned weights of the architecture prior ($\mu_a=0.56$, 0.75 and 0.09; the ``\nameref{sec:res:integrity}'' section) and the larger losses of every transfer method on tabular tasks: the main effects show not only which design choices transfer but also where transfer is weak.

With pools restricted to architectures below a parameter cap (1\,M parameters for convolutional networks and text encoders, 0.5\,M for tabular networks), QuaNAS remained the most efficient search (AUC 0.068 versus 0.134--0.185; Holm $p=0.0012$ against PTK-MR, QIEA-ZC and REA and $p=0.0015$ against RoBoT-style and LS; mean rank 1.75 of 14). Its capped final models satisfied all three light-model conditions (non-inferior accuracy, at most one third of the reference size, lower CPU latency) on 5 datasets, all of them image or time-series tasks (5 of 6), at 3--18\% of the reference size and 1.7--3.7 times lower latency. On text, the vocabulary embedding, which the cap does not constrain, dominates the size, and the fastText references are hundreds of times faster on a CPU; on tabular data the CatBoost references are already small and fast (0.05--0.06\,ms). The light-model criterion therefore holds where the references are large networks, on five of the six image and time-series datasets, but not where they are already small and fast (fastText, CatBoost).

\subsection*{Error analysis and failure cases}\label{sec:res:errors}

\textbf{Text topic classification.} On Yahoo Answers and 20 Newsgroups the final models were 3.4 and 4.2\,pp less accurate than fastText. The cause lies in the low-fidelity signal that every search shares: after the short training of two or five epochs, the bag-of-embeddings architectures of the text space, the family to which fastText belongs, ranked at best 166th and 282nd of 600, behind recurrent and convolutional encoders, whereas after full training the linear fastText models were the most accurate. Accordingly, the best architecture found by any of the four comparators whose architectures were trained was 3.1 and 5.7\,pp below fastText, while QuaNAS had the lowest regret on both tasks. The searched text models are also 300 to 1,000 times slower than fastText on one CPU thread (5.6--18.6\,ms versus 0.02--0.04\,ms per document). On the sentiment task Amazon Polarity, by contrast, QuaNAS exceeded the bigram fastText reference.

\textbf{OCTMNIST.} The official validation and test splits of OCTMNIST differ markedly: ResNet-18 reached 94.2\% validation but 78.5\% test accuracy. Model selection by validation accuracy, which every method used, cannot correct such a shift, and the QuaNAS model was 0.89\,pp below the reference, within the 77.0--78.3\% range of the comparators' architectures.

\textbf{Flat tabular landscapes.} Credit Default and Grid Stability were the hardest tasks for QuaNAS (AUC 0.216 and 0.146). Their best and median candidates differ by only 0.62 and 3.0\,pp, their stability coefficients are among the lowest (0.57 and 0.65) and no proxy is informative (best $\tau\approx0.06$). QuaNAS reached the near-optimal target within the budget in only 43\% and 23\% of runs there, yet still had the lowest regret of all primary comparators. On Credit Default, ranking by SynFlow alone happened to do better (0.194 versus 0.216), although SynFlow ranking was the worst method overall (mean AUC 0.355), a reminder that single proxies can win on single tasks but are not reliable across them.

\subsection*{Summary of the frozen hypotheses}\label{sec:res:hyp}

\Cref{tab:hyp} lists the seven hypotheses fixed before the confirmation, with their criteria and outcomes. Four were supported, two partly supported, and the operational check of the modality detector was met.

\begin{table}[!ht]
\centering
\caption{\textbf{Hypotheses of the frozen protocol, evaluated once on the 12 confirmation datasets.}}
\label{tab:hyp}
\footnotesize
\begin{tabularx}{\textwidth}{@{}l>{\raggedright\arraybackslash}X>{\raggedright\arraybackslash}X@{}}
\toprule
Hypothesis & Criterion fixed before the confirmation & Outcome \\
\midrule
Search efficiency & AUC lower than each primary comparator (Holm $p<0.05$) and best mean Friedman rank & Supported: Holm $p=0.0012$ for all five; mean rank 1.08 of 14\\
Speed & $\mathrm{TTT}_{0.05}$ lower than each primary comparator (Holm $p<0.05$) & Supported: Holm $p\le0.0068$\\
Final model & superior or non-inferior (margin 0.5\,pp) to the reference on at least 7 of 12 datasets & Supported: 9 of 12 (7 superior)\\
Components & lowest mean AUC among the full method and six removals, and no removal significantly better & Partly supported: no removal better (Holm $p=1$); fixed weights 0.056 versus 0.058\\
Light models & search rule in the capped pools; capped model non-inferior, at most one third of the reference size and faster on CPU on at least 7 of 12 & Partly supported: search rule met (Holm $p\le0.0015$); model criterion on 5 of 12\\
Zero-query transfer & prior optimum non-inferior to the reference on at least 7 of 12 and not worse than RoBoT-style transfer on at least 7 of 12 & Supported: 9 and 11 of 12\\
Modality detection (operational check) & 39 of 39 datasets and 6 of 6 synthetic cases & Met: 39 of 39 and 6 of 6\\
\bottomrule
\end{tabularx}
\end{table}

\section*{Discussion}\label{sec:discussion}

\textbf{Earlier searches are the knowledge that matters.} Three lines of evidence locate the efficiency of small-budget search in knowledge transferred from earlier searches. Removing the transferred prior or the prior-driven start raised regret significantly; on the held-out datasets every method that used the library, from the static ranking of a single nearest task to the methods that also learn from the new task's evaluations, had a lower mean regret than every pre-specified comparator, whereas the same GP without the library had more than twice the mean regret of QuaNAS; and the prior optimum alone, without a single evaluation, matched or exceeded the reference models on 9 of 12 datasets. With libraries of only 6 to 15 earlier pools, this knowledge was available for every modality studied, and, except on tabular data, it did not depend on near-duplicate datasets (see ``\nameref{sec:res:robust}''). Zero-cost proxies, in contrast, contributed little on average, in line with their weak and task-dependent correlations, and their computation adds to the cost of every search before its first query.

\textbf{Two complementary ways of reusing earlier searches.} Per-task surrogates, as in RGPE, keep a full model of every earlier task and weight these models by how well they rank the new task's evaluations; QuaNAS compresses every earlier task into additive main effects, pools them, checks by leave-one-task-out stacking whether the library predicts its own members, and lets the GP re-estimate how much the pooled prior and the proxies are trusted. Per-task surrogates retain interactions between decisions and pay off when an earlier task is a close match, as on NAS-Bench-201, where every earlier task contains the same candidates in almost the same order, and when only one or two earlier searches exist. Pooled main effects are coarser but degrade more gracefully when the earlier tasks are weak guides, as on the tabular tasks, where QuaNAS was better on 9 of 13 pools, and they turn the library into explicit knowledge: which option of each design decision transferred and how consistently the library supports it (\cref{fig:effects}). In practice, main-effect transfer suits weakly transferable spaces and settings in which the reused knowledge must be inspected, whereas per-task surrogates suit libraries of closely related tasks; a library from another kind of data in the same space is recognized only by a check against the new task's evaluations (\nameref{S1_Appendix}). Both mechanisms are cheap compared with meta-trained transfer methods such as MetaD2A, TNAS or KEGNAS~\cite{lee2021metad2a,shala2023tnas,wang2025kegnas}, which need large collections of earlier tasks to train dataset encoders, and post-hoc combinations of the two did not dominate either of them (\nameref{S1_Appendix}).

\textbf{Relation to proxy-based search.} Proxy-calibrating searches learn proxy weights from scratch; RoBoT-style spends half of its budget on optimizing them, which is consistent with its slow start (mean regret 0.215 after 25 queries) and its competitive end point (0.062 after 100 queries). On the three tabular datasets, the searched networks were more accurate than every boosted-tree baseline, and a plain two-layer MLP also beat the trees on two of them, so these datasets do not reproduce the advantage of tree ensembles reported for larger tabular benchmarks~\cite{grinsztajn2022tree}.

\textbf{Scope of application.} QuaNAS is most effective when a library of earlier searches in the same search space exists and the low-fidelity landscape has structure: on the image, time-series and text tasks it found near-optimal architectures after 1 to 37 queries on average. Flat, noisy landscapes are hard for every method, and, like every low-fidelity search, QuaNAS is misled when short training ranks architecture families differently from full training, as on the topic-classification tasks. A library that is coherent but stems from another kind of data in the same search space passes the leave-one-task-out check and can mislead the transfer (\nameref{S1_Appendix}), and a single earlier search cannot be checked at all, in which case QuaNAS falls back to the proxies. A new search space therefore needs a few earlier searches before its knowledge can be transferred; until then, the reliability switch and the GP make QuaNAS a proxy-guided Bayesian search.

\textbf{Practical implications.} For a practitioner with a library of earlier searches, transfer-based search that also learns from the new task's evaluations reached a near-optimal architecture after 2.4 to 3.3 minutes of short training on average, instead of 10 to 18 minutes for the proxy-based and evolutionary comparators, and QuaNAS needs neither labels of the new task for its transfer nor domain labels. On image and time-series data the final QuaNAS models were also smaller than standard references and, except on FetalECG, faster on a CPU; on text, latency-critical applications are better served by fastText.

\textbf{Limitations of the evidence.} Searches were run over fixed pools of real candidates: uniform samples of the convolutional space, stratified samples of the text space and the whole tabular space. Regret is measured against the best pool member and on the short-training objective; full training was affordable only for shortlists and selected architectures, and the model-level comparison between search methods is not symmetric. The reference models share the training recipe of the searched models and are not tuned to state-of-the-art accuracy. RoBoT was re-implemented on these pools with the code shared by all proxy-based comparators, the seven proxies were fixed before newer ones such as MeCo~\cite{jiang2023meco}, SWAP~\cite{peng2024swap} and AZ-NAS~\cite{lee2024aznas} were considered, and the comparisons with transfer-based methods were added after the confirmation, although with settings fixed before they were run. The library stress tests cover the convolutional space only, and the combinations of the two transfer mechanisms were explored post hoc on data that had been seen. Twelve confirmation datasets give limited power to resolve small differences in task-level tests, such as those between transfer methods. The protocol was frozen in our own version control rather than in a public registry, and transfer across search spaces is outside the scope of this study.

\section*{Conclusion}\label{sec:conclusion}

This paper introduced QuaNAS, a neural architecture search method that reuses earlier searches of the same search space as target-label-free priors and re-estimates, from every evaluation of the new task, how far these priors and zero-cost proxies are trusted. On 12 new datasets from four modalities and under a protocol frozen before any confirmation data were seen, QuaNAS lowered anytime regret 2.7- to 3.9-fold relative to five pre-specified baselines that start from scratch, reached the low-fidelity target three to five times sooner, and produced final models that matched or exceeded matched-recipe references on nine datasets, mostly at a fraction of their size. Even its prior alone, before any evaluation, selected architectures that matched or exceeded the references on nine datasets.

The post-hoc analyses sharpen what these results mean. On the held-out datasets every method that used the library of earlier searches had a lower mean regret than every method that did not, and zero-cost proxies contributed little and cost time. QuaNAS matched the strongest transfer method from hyperparameter optimization, RGPE, in mean regret (mean difference 0.000, 95\% CI $[-0.012, 0.013]$) with no negative transfer observed, while RGPE had the lower median regret and reached the target sooner. The two ways of reusing earlier searches proved complementary: across 47 pools and in stress tests, pooled main effects transferred better on weakly transferable tabular tasks and when the library carried no information, per-task surrogates on image, time-series and text tasks, with small or mismatched libraries and on NAS-Bench-201. The results therefore support three conclusions: with tasks rather than seeds as replicates, earlier searches are the knowledge that matters for small-budget NAS; target-label-free main-effect transfer with explicit uncertainty about its weight is a simple and inspectable way to reuse them, on par with the strongest transfer method in mean regret and the better of the two where transfer is weak; and reliable transfer requires checking the library against the new task, not only against itself. These conclusions hold for searches over candidate pools within shared search spaces and for budgets of about one hundred short trainings. Future work will add a target-level check of the library to QuaNAS, extend the transfer across search spaces, and correct the low-fidelity signal for architecture families that converge slowly under short training.

\section*{Supporting information}

\paragraph*{S1 Appendix.}
\label{S1_Appendix}
\textbf{Library stress tests and combinations of the two transfer mechanisms.} Post-hoc analyses on the 47 labeled pools: manipulated libraries (Table A) and combinations of per-task surrogates with the QuaNAS prior (Table B).

\section*{Data availability}
All data and code underlying the findings are fully available without restriction. All 39 datasets and the NAS-Bench-201 benchmark are public; the 27 development datasets are listed in the section "Development and confirmation datasets" and the 12 confirmation datasets in Table 3. The code, configurations, frozen-protocol documents with their freeze dates, the 47 candidate pools (short-training accuracies, proxy scores, costs), the per-run search, post-hoc, full-training and latency results, and the scripts behind every table and figure are archived on Zenodo at \url{https://doi.org/10.5281/zenodo.23562488}.

\section*{Funding}
This research is funded by the Indonesian Endowment Fund for Education (LPDP) on behalf of the Indonesian Ministry of Higher Education, Science, and Technology, and managed under the EQUITY Program (Contract No. 4299/B3/DT.03.08/2025 and No. 3029/PKS/ITS/2025), Directorate of Research and Community Service Institut Teknologi Sepuluh Nopember Ministry of Higher Education, Science, and Technology in accordance with the Implementation Agreement of the Strategic Research Grant (SRG) Assigned Research Program Institut Teknologi Sepuluh Nopember Fund for the Year 2026 Researcher Contract Number: 1669/PKS/ITS/2026, dated February 10, 2026. The funders had no role in study design, data collection and analysis, decision to publish, or preparation of the manuscript.

\section*{Competing interests}
The authors have declared that no competing interests exist.

\nolinenumbers


\clearpage
\section*{S1 Appendix. Library stress tests and combinations of the two transfer mechanisms}
\setcounter{table}{0}
\renewcommand{\thetable}{\Alph{table}}
\subsection*{Library stress tests}
Both analyses below are post hoc and use the labeled pools of the main paper (sections ``Experimental protocol'', ``Comparison with transfer-based Bayesian optimization'' and ``Robustness: library composition, manipulated libraries and a full search space''). AUC is the mean normalized anytime regret over 100 queries (lower is better). Negative transfer: a pool on which a method that uses the library of earlier searches has an AUC more than 0.01 above that of the same Gaussian process without the library (GP-BO). All methods use the Gaussian process of QuaNAS; RGPE, WS-GP and NT are defined in the section ``Compared methods'' of the paper. Code and per-run results: \texttt{posthoc/mismatch.py}, \texttt{posthoc/small\_lib.py} and \texttt{posthoc/hyb\_run*.py} in the code and data archive cited in the Data Availability statement of the article.

To probe the limits of transfer, we manipulated the library of the 21 convolutional pools (6 confirmation and 15 development pools, 10 seeds each; Table~\ref{tab:s1}) in three ways. Two are realistic: a library of only one or two earlier searches, drawn at random (five draws each), and a library of the other data type only (time-series pools for the 15 image targets and image pools for the 6 time-series targets). The third is extreme: the accuracies of every library task randomly permuted. RGPE coped better with the realistic perturbations. With one earlier search, the leave-one-task-out check of QuaNAS has nothing to compare, so the reliability switch fired in every run and QuaNAS fell back to a proxy-guided search (mean AUC 0.103 versus 0.080 for RGPE); with two earlier searches it used the transferred prior in most runs but remained behind (0.073 versus 0.055; two-sided $p=0.005$). A library of the other data type is consistent within itself (leave-one-task-out weight $\mu_a=0.36$ for the time-series library and 0.68 for the image library), so the switch did not fire and the transferred main effects misled QuaNAS (0.108; negative transfer on 6 of 21 pools), whereas the ranking weights of RGPE detected the mismatch from the new task's evaluations (0.061; one pool; $p=0.005$). Only when the library carried no information did the switch pay off: it fired on all 21 pools, and QuaNAS (0.120; 5 pools) lost less than RGPE (0.164; 12 pools; $p=0.004$), which cannot recognize an uninformative library before it has evaluated the new task. On the six confirmation pools alone the mean AUC differed in the same directions, but the counts of negative transfer were equal (Table~\ref{tab:s1}, lower block).

\begin{table}[!ht]
\centering
\caption{\textbf{Library stress tests on the 21 convolutional pools.}}
\label{tab:s1}
\small
\begin{tabular}{@{}lcccc@{}}
\toprule
Library & QuaNAS & RGPE & WS-GP & NT \\
\midrule
\multicolumn{5}{@{}l}{\emph{All 21 pools}}\\
Full library & 0.043 (1) & \textbf{0.034} (0) & 0.064 (0) & 0.083 (3)\\
Other data type only & 0.108 (6) & \textbf{0.061} (1) & 0.089 (3) & 0.235 (15)\\
Scrambled labels & \textbf{0.120} (5) & 0.164 (12) & 0.141 (9) & 0.279 (16)\\
One random earlier search & 0.103 (3) & \textbf{0.080} (0) & 0.088 (0) & 0.120 (5)\\
Two random earlier searches & 0.073 (2) & \textbf{0.055} (0) & 0.080 (1) & 0.093 (3)\\
\midrule
\multicolumn{5}{@{}l}{\emph{6 confirmation pools}}\\
Full library & 0.024 (0) & \textbf{0.023} (0) & 0.062 (0) & 0.074 (1)\\
Other data type only & 0.074 (1) & \textbf{0.051} (1) & 0.120 (2) & 0.243 (4)\\
Scrambled labels & \textbf{0.122} (3) & 0.141 (3) & 0.141 (5) & 0.283 (5)\\
One random earlier search & 0.083 (0) & \textbf{0.067} (0) & 0.085 (0) & 0.109 (3)\\
Two random earlier searches & 0.055 (0) & \textbf{0.029} (0) & 0.102 (1) & 0.059 (1)\\
\bottomrule
\end{tabular}
\begin{flushleft}\footnotesize Mean AUC (10 runs per pool; lowest per row in bold) and, in parentheses, the number of pools with negative transfer (GP-BO without library: mean 0.147 over the 21 pools). Other data type only: time-series library pools for image targets and image pools for time-series targets. Scrambled labels: the accuracies of every library task randomly permuted. One or two random earlier searches: five random library subsets of that size, two seeds each.
\end{flushleft}
\end{table}

\subsection*{Combinations of the two transfer mechanisms}
After the comparison with RGPE we tried to obtain the advantages of both mechanisms. Eleven combinations were evaluated on all 47 pools and on NAS-Bench-201 (Table~\ref{tab:s2}): the GP surrogates of the earlier tasks as additional priors with random-effect weights and leave-one-task-out stacking (with cross-fitting when the new pool shares fewer than half of its architectures with the library pools), variants with an exploration scale chosen on the library, and a choice between QuaNAS and such surrogate priors by simulated searches on the library. Two further variants used the QuaNAS GP as the target model of RGPE; screened on eight pools (five confirmation and three development pools), they reached a mean AUC of 0.145 and 0.142, against 0.094 for QuaNAS and 0.157 for RGPE on the same pools, and were not pursued (not shown in Table~\ref{tab:s2}). A selection rule was fixed before any of them was run: a lower mean, median and 90th percentile of AUC over the 47 pools than both QuaNAS and RGPE, and a NAS-Bench-201 AUC within 0.01 of RGPE. None had a lower mean, median and 90th percentile of AUC than both QuaNAS and RGPE (best mean 0.074, against 0.070 and 0.079). Even an oracle that picked the better of the two for every pool in hindsight would lower the mean AUC to 0.056 but leave the median of RGPE unchanged (0.030): the two mechanisms win on different tasks rather than one containing the other.

\begin{table}[!ht]
\centering
\caption{\textbf{Combinations of per-task surrogates and the QuaNAS prior (47 pools; NAS-Bench-201).}}
\label{tab:s2}
\small
\begin{tabular}{@{}lcccc@{}}
\toprule
Method or variant & Mean AUC & Median & 90th pct. & NAS-Bench-201 \\
\midrule
QuaNAS & 0.070 & 0.036 & 0.172 & 0.049\\
RGPE & 0.079 & 0.030 & 0.211 & 0.002\\
Oracle: better of the two per pool & 0.056 & 0.030 & 0.149 & --\\
\midrule
surrogate + main-effect prior, per-task random effects & 0.084 & 0.034 & 0.260 & 0.008\\
surrogate prior only & 0.091 & 0.039 & 0.329 & 0.007\\
one random effect on the surrogate mean & 0.076 & 0.037 & 0.197 & 0.004\\
no random effects on the surrogates & 0.080 & 0.041 & 0.228 & 0.004\\
stacking without cross-fitting\textsuperscript{\dag} & 0.084 & 0.034 & 0.260 & 0.008\\
stacking always cross-fitted & 0.085 & 0.041 & 0.283 & 0.008\\
proxy prior always used & 0.085 & 0.038 & 0.252 & 0.005\\
proxy prior never used\textsuperscript{\dag} & 0.084 & 0.034 & 0.260 & 0.008\\
exploration scale chosen on the library & 0.083 & 0.037 & 0.262 & 0.006\\
QuaNAS or surrogate prior, chosen on the library & 0.074 & 0.037 & 0.221 & 0.006\\
as above, one random effect on the surrogate mean & 0.076 & 0.034 & 0.200 & 0.005\\
\bottomrule
\end{tabular}
\begin{flushleft}\footnotesize Per-pool mean AUC (confirmation pools 30 seeds, development pools 10 seeds); mean, median and 90th percentile over the 47 pools; NAS-Bench-201: mean over its three datasets. Variants: GP surrogates of the earlier tasks (the base models of RGPE) combined with the QuaNAS prior inside its Gaussian process. \textsuperscript{\dag}Differs from the first variant only on the pools on which that variant cross-fitted its stacking (9 pools) or fired the reliability switch (2 development pools); the per-pool results differ on 7 and 2 pools, respectively, but the summary statistics coincide at three decimals (mean AUC 0.0839, 0.0844 and 0.0838 for the first variant and the two marked rows), and neither mechanism was triggered on NAS-Bench-201.
\end{flushleft}
\end{table}

\end{document}